\documentclass[lettersize,journal]{IEEEtran}

\usepackage{amsmath,amsfonts}
\usepackage{algorithmic}
\usepackage{algorithm}
\usepackage{array}
\usepackage{subcaption}
\usepackage{textcomp}
\usepackage{stfloats}
\usepackage{url}
\usepackage{verbatim}
\usepackage{graphicx}
\usepackage{cite}
\usepackage{xspace}
\usepackage[dvipsnames]{xcolor}
\usepackage{lipsum}
\usepackage{multirow}
\usepackage[english]{babel}
\usepackage{hyperref}
\usepackage{cleveref}
\usepackage{siunitx}
\usepackage{acronym}
\usepackage{standalone}
\usepackage{tikz}
\usepackage{bm}
\usepackage{booktabs}
\usepackage{gensymb}

\newcommand{\LC}[1]{\textcolor{red}{#1}}
\newcommand{\omitted}[1]{}

\usetikzlibrary{positioning}
\usetikzlibrary{shapes}

\usepackage{hyperref}
\usepackage{tikz}
\usepackage{pgfplots}
\usepackage{siunitx}
\usepgfplotslibrary{groupplots}
\usetikzlibrary{backgrounds,arrows,automata,shapes,positioning,calc,through}
\tikzset{
  arrow/.style={
    thick,
    ->,
    >=stealth},
  darrow/.style={
    thick,
    <->,
    >=stealth},
  block/.style={
    draw,
    rectangle,
    rounded corners,
    inner sep=0pt
  }
}

\hypersetup{colorlinks=true}

\acrodef{IMU}{Inertial Measurement Unit}
\acrodef{ICP}{Iterative Closest Point}
\acrodef{SLAM}{Simultaneous Localization And Mapping}
\acrodef{DARPA}{Defense Advanced Research Projects Agency}
\acrodef{SubT}{Subterranean}
\acrodef{MAD}{Median of Absolute Deviation}
\acrodef{UGVs}{Unmanned Ground Vehicles}
\acrodef{UAVs}{Unmanned Aerial Vehicles}
\acrodef{GNSS}{Global Navigation Satellite System}
\acrodef{Norlab}{Northern Robotics Laboratory}

\DeclareSIUnit\cpu{CPU}

\usepackage{color}
\definecolor{dark_red}{rgb}{0.4, 0.0, 0.0}
\definecolor{light_red}{rgb}{0.8, 0.1, 0.1}
\definecolor{dark_green}{rgb}{0.0, 0.4, 0.0}
\definecolor{light_green}{rgb}{0.1, 0.8, 0.1}
\definecolor{dark_blue}{rgb}{0.0, 0.0, 0.4}
\definecolor{light_blue}{rgb}{0.1, 0.1, 0.8}
\definecolor{dark_violet}{rgb}{0.4, 0.1, 0.4}
\definecolor{light_violet}{rgb}{0.8, 0.1, 0.8}
\definecolor{dark_orange}{rgb}{0.4, 0.4, 0.1}
\definecolor{light_orange}{rgb}{0.8, 0.6, 0.1}

\newcommand{\parlength}[1]{}
\newcommand{\myParagraph}[1]{{\bf #1.}}
\newcommand{\eg}{\emph{e.g.,}\xspace}
\newcommand{\ie}{\emph{i.e.,}\xspace}
\newcommand{\setal}{~\emph{et al.}\xspace}
\newcommand{\lidar}{LIDAR\xspace}
\newcommand{\lidars}{LIDARs\xspace}
\newcommand{\Lidar}{LIDAR\xspace}
\newcommand{\costar}{CoSTAR\xspace}
\newcommand{\ccn}{CTU-CRAS-Norlab\xspace}

\newcommand{\edited}[1]{#1}

\newcommand{\SubT}{SubT\xspace}

\usepackage{subfiles}
\graphicspath{{figures/}}

\usepackage{ifthen}
\newboolean{printBibInSubfiles}
\setboolean{printBibInSubfiles}{true} 
\def\bib{\ifthenelse{\boolean{printBibInSubfiles}}
        {\bibliographystyle{IEEEtran}\bibliography{bibliography.bib}}
        {}
    } 

\begin{document}
\setboolean{printBibInSubfiles}{false}
\title{Present and Future of SLAM in Extreme Underground Environments}

\author{Kamak Ebadi, 
Lukas Bernreiter,
Harel Biggie,
Gavin Catt,
Yun Chang,
Arghya Chatterjee,
Christopher E. Denniston,
Simon-Pierre Deschênes,
Kyle Harlow, 
Shehryar Khattak,
Lucas Nogueira, 
Matteo Palieri,
Pavel Petr\'{a}\v{c}ek,
Mat\v{e}j Petrl\'{i}k, 
Andrzej Reinke,
V\'{i}t Kr\'{a}tk\'{y}, 
Shibo Zhao, 
Ali-akbar Agha-mohammadi, 
Kostas Alexis, 
Christoffer Heckman, 
Kasra Khosoussi,
Navinda Kottege, 
Benjamin Morrell,
Marco Hutter,
Fred Pauling, 
François Pomerleau, 
Martin Saska,
Sebastian Scherer, 
Roland Siegwart,
Jason L. Williams,
Luca Carlone
\thanks {\edited{This work was supported by the Jet Propulsion Laboratory -- California Institute of Technology, under a contract with the National Aeronautics and Space Administration. © 2022. All rights reserved. This work was partially supported by the Defense Advanced Research Projects Agency (DARPA) under Agreement No. HR00111820045. 
The presented content and ideas are solely those of the authors.}}
}

\maketitle


\begin{abstract}
This paper surveys recent progress and discusses future opportunities for Simultaneous Localization And Mapping (SLAM) in extreme underground environments.
SLAM in subterranean environments, \edited{from tunnels, caves, and man-made underground structures on Earth, to lava tubes on Mars}, is a key enabler for a range of applications, such as planetary exploration, search and rescue, disaster response, and automated mining, among others. SLAM in  underground environments has recently received substantial attention, thanks to the \emph{DARPA Subterranean (\SubT) Challenge}, a global robotics competition aimed at assessing and pushing the \edited{state of the art} in autonomous robotic exploration and mapping in complex underground environments.
This paper reports on the \edited{state of the art} in underground SLAM by discussing different SLAM strategies and results across six teams that participated in the three-year-long \SubT competition. 
In particular, the paper has four main goals. First, we review the algorithms, architectures, and systems adopted by the teams; particular emphasis is put on \lidar-centric SLAM solutions 
\edited{(the go-to approach for virtually all teams in the competition)}, heterogeneous multi-robot operation \edited{(including both aerial and ground robots)}, and real-world underground operation (from the presence of obscurants to the need to handle tight computational constraints).
We do not shy away from discussing the ``dirty details'' behind the different \SubT SLAM systems, which are often omitted from technical papers. 
Second, we discuss the maturity of the field by highlighting what is possible with the current SLAM systems and what we believe is within reach with some good systems engineering. 
Third, we outline what we believe are fundamental open problems, that are likely to require further research to 
break through.  
Finally, we provide a list of open-source SLAM implementations and datasets that have been produced during the \SubT challenge and related efforts, and constitute a useful resource for researchers and practitioners.
\end{abstract}


\section{Introduction\parlength{(1.5 pages)}}

Simultaneous Localization and Mapping (SLAM) remains at the center stage of robotics research, 
after more than 30 years \edited{since its inception}. 
SLAM is without a doubt a mature field of research, and the advances over the last three decades keep steadily transitioning into industrial applications, from domestic robotics~\cite{Roomba, Astro, Dyson}, to self-driving cars~\cite{SLAM-selfdriving} and virtual and augmented reality goggles~\cite{Hololens, Oculus}. At the same time, its pervasive nature and its blurry boundaries as a robotics subfield still leave space for exciting research progress. While previous survey efforts have targeted SLAM in general~\cite{cadena2016slam}, SLAM is also actively investigated in specific subdomains, from deployment on nano-drones~\cite{Navion} to city-scale mapping~\cite{AutoMerge}, to deployment in perceptually challenging conditions. This paper surveys algorithms and systems for \lidar-centric SLAM in extreme underground environments.

\myParagraph{Present and Future of SLAM in Underground Worlds}
The past two decades have seen a growing demand for autonomous exploration and mapping of diverse subterranean environments, from tunnels and urban underground environments, to complex cave networks. This has led to an increasing attention towards underground SLAM, which is a key enabler for navigation in GPS-denied environments where a-priori maps are unavailable. 
Mature SLAM systems for subterranean mapping have the potential to enable a range of terrestrial and planetary applications, from surveying, search and rescue, disaster response, and automated mining, to exploration of planetary caverns that could hold clues about the evolution and habitability of the early Solar System. 

Progress in underground SLAM has been particularly catalyzed by the recent \emph{DARPA Subterranean (\SubT)
 Challenge}~\cite{subt_challenge}, a three-year-long global competition ended in 2021, and having the goal of demonstrating and advancing the \edited{state of the art} in mapping and exploration of complex underground environments.
 \edited{The competition had a \emph{systems} track and a \emph{virtual} track, and included three main events: the Tunnel circuit event, the Urban circuit event, and the Finals.}
The teams competing in the systems track had the goal of deploying a team of robots to map a sequence of large-scale, unknown  underground environments (including caves, tunnels, and subways), detect \emph{artifacts} (\ie objects of interest, including survivors, mobile phones, \edited{fire extinguishers, etc.}), and report their locations with stringent performance requirements (\ie within \SI{5}{\metre} errors, in \edited{underground} networks branching for hundreds of meters to kilometers). While the team of robots was supervised by a single human operator, communication constraints as well as the fast-pace of the competition (the robots had to complete the exploration in under 1 hour) pushed the teams to develop robust and highly autonomous solutions that required minimal human intervention. 

\myParagraph{Technical Challenges for Underground SLAM}
Robots exploring underground environments typically do not have access to sources of absolute positioning (\eg GNSS) and rarely have access to prior maps of the environment.
\edited{While in many cases (e.g., search and rescue operations) building a map {is not} the goal of the deployment, mapping remains a crucial prerequisite for successful underground operation. Mapping these environments is particularly challenging; poor} lighting conditions make it challenging to deploy  visual and visual-inertial SLAM solutions; while the lack of illumination can be partially compensated by onboard light sources, the resulting illumination is either tenuous or creates specular reflections that interfere with visual feature tracking. Beyond cameras, other sensors are also challenged by the strenuous conditions found in the undergrounds. The potential presence of dense obscurants, such as  fog, whirling dust clouds, and smoke, challenges the use of \lidars. The use of fast-moving platforms on rough terrains induces noise in inertial sensors, due to the aggressive 6-DoF motion and high-frequency vibrations. 

Even when the sensors themselves perform to specifications, these environments create further challenges for SLAM algorithms. For instance, the lack of perceptual features \edited{(\eg long corridors, large open spaces, and chambers)} induces failures in \lidar-odometry approaches based on feature or scan matching. Similarly, the presence of self-similar and symmetric areas, 
\edited{and the lack of distinctive visual texture} 
increase the number of false positives in the place recognition methods that fuel loop closure detection in SLAM, \edited{and map fusion and merging in multi-robot systems}. The complex and ambiguous terrain topography is further exacerbated by sudden changes in the scale of the environment (\eg a small tunnel leading to a large cave), which clashes with potential scenario-dependent parameter tuning in SLAM systems. 

The challenges of underground SLAM extends to system engineering. SLAM algorithms must operate on-board under computational constraints, which are particularly stringent on aerial platforms, and also require careful parameter tuning and code optimizations on wheeled and legged robots. Moreover, these SLAM systems are required to withstand  intermittent and faulty sensor measurements, as well as unexpected motions and shocks due to potential robot falls and collisions.

\myParagraph{Related Surveys}
Progress in SLAM research has been reviewed by Durrant–Whyte and Bailey \cite{survey1, survey2} and more recently by Cadena\setal~\cite{cadena2016slam}.
Other relevant surveys have recently focused on multi-robot SLAM and related applications.
Kegeleirs\setal~\cite{survey3} and Dorigo\setal~\cite{survey6} provide an overview of challenges in SLAM with robotic swarms and their application for gathering, sharing, and retrieving information.
Halsted\setal~\cite{survey4} survey distributed optimization algorithms for multi-robot applications. 
Parker\setal~\cite{survey5} examine multi-robot SLAM architectures with focus on communication issues and their impact on multi-robot teams.
Lajoie\setal~\cite{survey7} provide a literature review of collaborative SLAM with focus on  robustness, communication, and resource management.
Zhou\setal~\cite{survey8} review algorithmic developments in making multi-robot systems robust to environmental uncertainties, failures, and adversarial attacks.
Prorok\setal~\cite{Prorok22axiv-resilience} discuss resilience in multi-robot systems. None of these surveys focus on SLAM in underground environments. 

\myParagraph{Contribution}
This paper reports on the state of the art and state of practice in underground SLAM by discussing different SLAM strategies and results across six teams that participated in the three-year-long \SubT challenge. 
In particular, the paper has four main goals. First, we provide a broad review of related work (\Cref{sec:relatedWork})
and then delve into the single- and multi-robot  SLAM architectures  adopted by six teams that participated in the systems track of the DARPA SubT challenge (\Cref{sec:architectures}); particular emphasis is put on multi-modal \lidar-centric SLAM solutions, heterogeneous multi-robot operation, and real-world underground operation. We also discuss the ``dirty details'' behind the different \SubT SLAM systems, which are often omitted from technical papers.
Second, we discuss the maturity of the field by highlighting what is possible with the current SLAM systems and what we believe is within reach with some good system engineering (\Cref{sec:maturity}). 
Third, we outline what we believe are fundamental open problems, that are likely to require further research to break through (\Cref{sec:openProblems}).  
Finally, we provide a list of open-source SLAM implementations and datasets that have been produced during the \SubT challenge and related efforts, and constitute a useful resource for researchers and practitioners. 
These are summarized in~\Cref{tab:code_and_dataset}.

\bib


%
\section{Overview of Related Work}
\label{sec:relatedWork}

This section provides a brief overview of related work on SLAM \emph{systems} for subterranean 
environments and multi-robot teams, 
before delving into the details of modern systems in~\Cref{sec:architectures}. {Early efforts on SLAM in subterranean environments} trace back to the work of Thrun\setal~\cite{thrun2003system} and Nuchter\setal~\cite{nuchter20046d}, which highlighted the importance of underground mapping and introduced early solutions involving a cart pushed by a human operator, or teleoperated robots equipped with laser range finders to acquire volumetric maps of underground mines.
Tardioli\setal~\cite{tardioli2012robot, tardioli2017robotized} present a SLAM system for exploration of underground tunnels using a team of robots. The system comprised of a navigation control module, a feature-based robot localization module, a communication module, and a supervisor module for multi-robot collaborative exploration in a tunnel. Zlot\setal~\cite{zlot2014efficient} propose a 3D SLAM system consisting of a 2D spinning lidar and an industrial-grade MEMS IMU to map over \SI{17}{\km}  of an underground mine. 
Kohlbrecher\setal~\cite{kohlbrecher2011flexible} present Hector SLAM, a flexible and scalable SLAM system with full 3D motion estimation developed specifically for urban search and rescue. The system consists of a navigation filter that uses an IMU for attitude estimation, and a  2D SLAM system for position and heading estimation within the ground plane.

Lajoie\setal~\cite{lajoie2020door} present DOOR-SLAM, a multi-robot SLAM system which consists of two key modules, a pose graph optimizer (combined with a distributed pairwise consistent measurement set maximization algorithm to reject spurious inter-robot loop closures), and a distributed SLAM front-end that detects inter-robot loop closures without exchanging raw sensor data. Chang, Tian,\setal~\cite{chang2021kimera,Tian21tro-KimeraMulti} present Kimera-Multi, a distributed multi-robot system for dense metric-semantic SLAM. Each robot builds a local trajectory estimate and a 3D mesh. When robots are within communication range, they initiate a distributed place recognition and robust pose graph optimization protocol based on graduated non-convexity. 

Autonomous exploration of extreme underground environments has received significant attention in the context of the { DARPA SubT Challenge}.
The competition gave rise to and inspired breakthrough technologies and capabilities in the field of underground SLAM \cite{ebadi2020lamp,palieri2020locus,ebadi2021dare,morrell2020robotic,2021LPICo2595.8122B,rouvcek2019darpa,petrlik2020robust,bouman2020autonomous,williams2020online,kramer2021vi,azpurna2021three,kratky2021autonomous,miller2020mine,ginting2021chord,azpurua2021towards,dang2020autonomous,wisth2021unified,khattak2020complementary,queralta2020collaborative,gross2019field,ohradzansky2021multi,bayer2019speeded,tidd2021passing,lindqvist2021compra,rogers2020test,carter2021darpa}. 
We review the details of key (multi-robot) SLAM systems developed in the context of the DARPA \SubT challenge in the next section.
\bib


\section{\edited{State of the Art} in Underground SLAM\parlength{(9 pages)}}
\label{sec:architectures}

This section examines the SLAM architectures adopted by six of the teams that participated in the systems track of the DARPA \SubT Challenge, and highlights the important design choices, differences, and common themes that enabled autonomous exploration of unknown underground environments. Moreover, this section provides a table of open-source implementations and datasets that are made publicly available by each team.
In particular,~\Cref{sec:anatomy} reviews the standard architecture of a multi-robot SLAM system and provides basic terminology. 
\Cref{sec:team-CERBERUS} to \Cref{sec:team-MARBLE} describe the specific SLAM architectures adopted by the six \SubT teams and highlight key design choices and ``dirty details''.~\Cref{sec:common-themes} discusses common themes, and includes a table of open-source implementations and datasets \edited{(\Cref{tab:code_and_dataset})}.

\subsection{\edited{Anatomy of Single- and Multi-Robot SLAM Systems}}
\label{sec:anatomy}

The architecture of a SLAM system typically includes two main components: 
the front-end and back-end~\cite{cadena2016slam};

The {\bf SLAM front-end} is in charge of abstracting the raw sensor data into more compact intermediate representations (\eg odometry, loop closures, landmark observations). For instance, a \lidar-based SLAM front-end may process \lidar scans into odometry estimates either by registering salient features extracted from consecutive \lidar scans ---an approach adopted by teams CERBERUS (Section~\ref{sec:team-CERBERUS})  and Explorer (Section~\ref{sec:team-explorer})--- or by dense registration of \lidar point clouds (or surfels) using ICP or its variants ---as adopted by teams \costar (Section~\ref{sec:team-CoSTAR}), CSIRO (Section~\ref{sec:team-CSIRO}), \ccn (Section~\ref{sec:team-ccn}) and MARBLE (Section~\ref{sec:team-MARBLE}). 

The {\bf SLAM back-end} is in charge of building robot trajectory and map estimates by fusing the intermediate representations produced by the front-end. The back-end typically includes a nonlinear estimator, with the de-facto standard approach being maximum a-posteriori estimation via \emph{factor graph optimization}~\cite{cadena2016slam}; this indeed has been adopted by virtually all teams below. A popular instance of factor graph optimization is \emph{\edited{blue}{pose graph} optimization}, where one optimizes the robot trajectory using relative pose measurements. The SLAM back-end can perform tightly-coupled and loosely-coupled sensor fusion, where the former fuses fine-grained measurements by different sensors (\eg 2D image features and inertial data), while the latter fuses intermediate estimates (\eg relative poses produced by a \lidar and camera). 
Tightly-coupled approaches are generally more accurate, as they rely on more precise models of the sensor data and its noise. \edited{Loosely-coupled approaches are easier to implement (\ie they are more modular) and often more convenient (\eg they give access to standard tools for outlier-rejection and health monitoring~\cite{yang2020graduated,santamaria2019towards}), but at the cost of decreased accuracy.}

{\bf Multi-robot SLAM} systems are characterized by the fact that sensor data is simultaneously collected by multiple robots, which are in charge of building a consistent map of the environment.
Multi-robot SLAM architectures can be \emph{centralized}, \emph{decentralized}, or \emph{distributed}. 
In centralized architectures, a base station collects data from all the robots (\eg raw sensor data or intermediate representations from the single robot front-ends) and then computes optimal trajectory and map estimates for the entire team. 
Each robot typically runs a local SLAM front-end (and possibly a local back-end) to pre-process the sensor data --- this reduces the amount of data to be transmitted and the subsequent computation at the base station; then, the base station may implement a \emph{multi-robot front-end}, which is in charge of detecting inter-robot loop closures, and a \emph{multi-robot back-end}, that estimates the robots' trajectories and map. 
\edited{In this paper, we call a multi-robot architecture \emph{decentralized} if each robot is treated as a base station: it collects all the data from the other robots and performs joint estimation of the trajectory and global map of the entire team.
Finally, we call an architecture \emph{distributed} if each robot only exchanges partial information with its neighbors and only estimates its own map by relying on distributed inter-robot loop closure detection and distributed optimization protocols~\cite{Tian21tro-KimeraMulti, tian2021resource, giamou2018talk, tian2018near, tian2021distributed}.
}
The following sections describe the SLAM architectures for each \SubT team.

\bib


\begin{figure*}[t]
\centering
	\includegraphics[width=0.95\textwidth, trim= 0mm 0mm 0mm 0mm, clip]{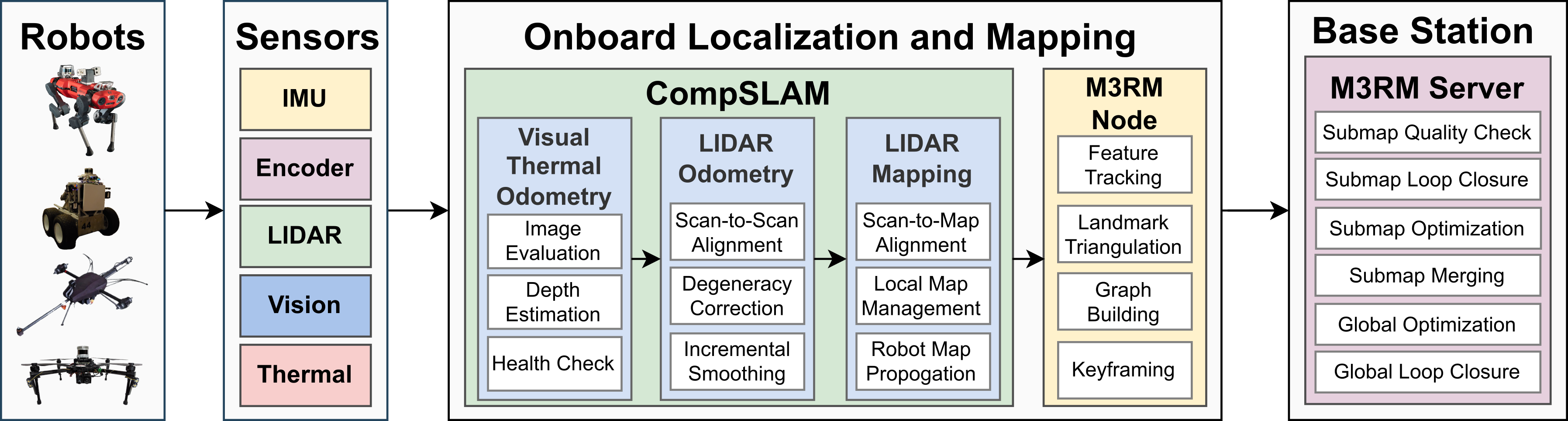}
	\caption{{Overview of team CERBERUS' SLAM architecture. Each robot estimates and operates on its own individual map. Periodically, these maps are sent to the mapping server on the base station for accumulation and for global multi-robot optimization}.\label{fig:cerberus:overview}  \vspace{-3mm}}
\end{figure*}

\subsection{Team CERBERUS}
\label{sec:team-CERBERUS}

Team CERBERUS won the Final event of the DARPA \SubT Challenge; their SLAM architecture is given in~\Cref{fig:cerberus:overview}. The architecture is powered by~\emph{CompSLAM}~\cite{khattak2020complementary}, a complementary multi-modal odometry and local mapping approach running at each (walking, flying, or roving) robot, and~\emph{M3RM}, a multi-modal, {multi-robot mapping server running at the base station.}  

\myParagraph{Onboard Odometry and Mapping via CompSLAM}
CompSLAM~\cite{khattak2020complementary} is a loosely-coupled approach that allows hierarchical fusion of a set of sensor-specific pose estimators \edited{as each estimate is refined by the next estimator. This enables operating in parallel into a single odometry estimate, while performing data- and process-level health checks~\cite{vtio2019}.}
In particular, CompSLAM performs a coarse-to-fine fusion of independent pose estimates including visual, thermal, depth, inertial, and possibly kinematic odometry sources. 
This loosely-coupled methodology provides redundancy and ensures robustness against perceptually degraded conditions, including self-similar geometries, low-light and low-texture scenes, and obscurants-filled environments (\eg fog, dust, smoke), assuming that each condition only affects a subset of sensors.

The visual- and thermal-inertial fusion (VTIO) components of CompSLAM build upon the work~\cite{bloesch2015robust} and extends it to exploit $16\textrm-{bit}$ raw data from LongWave InfraRed cameras~\cite{khattak2019robust,ktio2020} and depth from \lidar. 
Furthermore, the depth data from the \lidar is utilized to initialize or improve depth estimates of features tracked in visual and thermal imagery, providing robustness for scale estimation without the need for computationally expensive stereo-matching. 

The \lidar Odometry And Mapping component of CompSLAM develops on top of LOAM~\cite{zhang2014loam}. This component, along with VTIO priors, utilizes \lidar point clouds to perform a \lidar Odometry (LO) scan-to-scan matching and scan-to-submap matching \lidar Mapping (LM) step. Accordingly, the robot estimates its pose in the map and simultaneously constructs a local map of the environment. 
Following the hierarchical fusion approach, the estimates of the LO module are utilized by and refined upon by the LM module. To assess the quality at each iterative optimization step, the system \edited{utilizes a threshold} on the eigenvalues of the underlying approximate Hessian~\cite{dellaert2017factor, ebadi2021dare}, to identify the degrees of freedom that are possibly ill-conditioned due to geometric self-similarity. In case certain directions are determined to be ill-conditioned, the pose estimates from the previous estimator in the hierarchy \edited{(e.g., visual-inertial odometry)} are propagated forward, skipping the ill-conditioned module.

Finally, to produce smooth and consistent pose estimates, CompSLAM uses a factor-graph-based fixed-lag smoother, \edited{implemented as part of the LO module}, with a smoothing horizon of $3$ seconds. The factor graph is implemented using GTSAM~\cite{dellaert2012factor} and integrates relative LO estimates with IMU pre-integration factors~\cite{Forster17tro}. 
 To reduce pose drift and improve IMU bias estimation, zero-velocity factors are added when more than one sensing modality reports no motion for $0.5$ (consecutive) seconds. Moreover, during periods of no motion, roll and pitch estimates ---calculated directly from bias-compensated IMU measurements--- are added as prior factors.

\myParagraph{Multi-Robot Mapping and Optimization (M3RM)}
The core component of the CERBERUS multi-modal and multi-robot mapping (M3RM) approach is a centralized mapping server that utilizes multiple modalities such as \edited{\lidar, vision, IMU, wheel encoders}, etc., in a single factor graph optimization.
The deployed M3RM approach is based on the existing framework \emph{maplab}~\cite{schneider2018maplab} and can generally be subdivided into two components, namely the \emph{M3RM node} and \emph{server}. 

The M3RM node runs onboard each robot and is in charge of creating a local factor graph capturing multi-sensor  data collected and pre-processed by the robot,~\eg~odometry factors from CompSLAM.
The node also tracks BRISK~\cite{leutenegger2011brisk} features and triangulates the features to a visual map using the CompSLAM pose estimates.
Additionally, the \lidar scans (as well as the corresponding timestamps and extrinsic calibration) are attached to 
the factor graph. The factor graph is broken into submaps. To reduce bandwidth, each \lidar scan is compressed \edited{using DRACO~\cite{DRACO}} before transmission, reaching a total size of approximately $\SI{2}{megabytes}$ per submap. When robots establish a connection to the base station, the M3RM node transmits the completed submaps \edited{to the M3RM server.}
A synchronization logic ensures that only a completed submap transmission will be integrated into the multi-robot map.

The M3RM server runs at the base station and is in-charge of keeping track of all individual submaps for each robot and integrating them into a globally consistent multi-robot map. 
\edited{During the mission, the M3RM server allows a human operator to visualize the individual maps as well as the globally optimized multi-robot map which enables mission planning.
Moreover, the server has certain management functions such as removal of maps, performance profiles, and allows switching between CompSLAM and M3RM map per robot. The CompSLAM maps are not attached to the global multi-robot map but can be visualized using an overlay.}
To integrate the individual robot submaps into a single multi-robot map, the M3RM server first processes each incoming submap using a set of operations, namely 
(i) visual landmark quality check, (ii) visual loop closure detection, (iii) \lidar registrations, and (iv) submap optimization.
Since each submap's processing is independent of the processing of other submaps, the mapping server can process up to four submaps in parallel.
For visual loop closure detection, the method presented in~\cite{lynen2015get} is performed using the tracked BRISK features and an inverted multi-index. Correctly identified visual loop closures within a submap are implemented by merging the corresponding landmarks and are then integrated during the submap optimization.
Moreover, additional \lidar constraints are added to the factor graph by aligning consecutive scans within a submap using ICP.
Since the onboard odometry and mapping pipeline already provides an estimate of the poses, a prior transformation is readily available for each registration. 
However, if the resulting transformation differs significantly from the prior, it is rejected for robustness reasons as we expect that the drift between consecutive nodes is relatively small. 

After individual submaps processing, they are merged into the global multi-robot map, which is continuously optimized by the M3RM server.
A predefined set of operations are executed in an endless loop on the global multi-robot map, 
\edited{i.e., (i) 
multi-robot visual loop closure detection, \edited{(ii) multi-robot \lidar registrations}, and (iii) factor graph optimization.
In this case, these operations are performed on the entire multi-robot map, and have the goal of 
detecting intra- and inter-robot loop closures and performing a joint optimization.}

\myParagraph{Dirty Details}
\edited{\textit{Parameter tuning:} As for many other systems reviewed in this paper}, the top performance of team CERBERUS' SLAM solution requires a careful fine-tuning of all available parameters.
For example, the degeneracy detection relies on a hand-tuned set of parameters and is robot-dependent. The tuning is performed by using a grid search over several clusters of parameters and measuring their performance across relevant environments.
To complicate things further, the configurable parameters for the M3RM server have to be consistently applied to all robots in the global multi-robot map, making fine-tuning for specific robot types (e.g., flying and legged systems) as well as sensors (e.g., various camera and LIDAR systems) difficult at the stage of global mapping.

\textit{Covariances:} While it is desirable to dynamically adjust the covariances in the factor graphs depending on the quality of the sensor data, it proved challenging to balance the uncertainty of the visual and \lidar factors; 
therefore, the system relied on static (manually tuned) covariances for the latter. 

\textit{Loop closures:} None of the deployed robots performed onboard loop closure detection. Thus, in scenarios where robots stay out of communication range from the base station for a considerable time, the CompSLAM errors may accumulate, making it harder for the M3RM server to correct the estimates. 
\edited{Moreover, an incorrect robot map can ``break'' the whole global multi-robot map, which is why a human operator is needed to monitor and possibly remove specific robots from the multi-robot map.}
\omitted{{\bf This seems more of a limitation, rather than dirty details:}
Third, in CompSLAM, each odometry is assessed with an individual health metric that relates to the growth of the position covariance in visual-inertial methods, and to the numerical conditioning in the iterative nonlinear optimization steps during scan-to-scan matching. This implies that our method that does not assess the performance of each odometry source against another. Thus, although not often demonstrated in practice, it can be susceptible to tuning against specific environments. 
}

\subsection{Team CoSTAR}
\label{sec:team-CoSTAR}

\begin{figure*}[t!]
    \centering
    \includegraphics[trim={0cm 0cm 0cm 0cm},clip, width=1.0\textwidth]{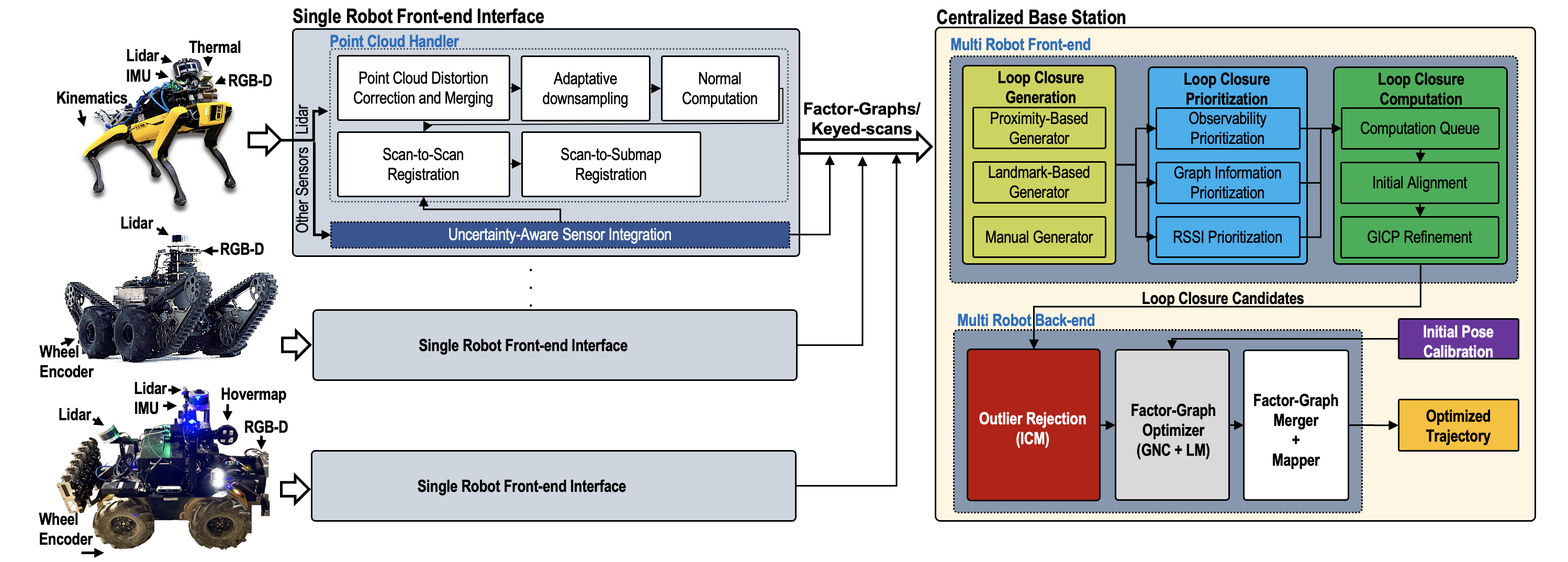}
    \caption{Overview of team CoSTAR's SLAM architecture (LAMP). Each robot runs a local front-end and communicates to the base station, which runs a multi-robot front-end (for loop closure detection) and back-end (for pose graph optimization). \label{fig:lamp_architecture} \vspace{-3mm}}
\end{figure*}

Team CoSTAR won the Urban event of the DARPA \SubT Challenge.
An overview of team CoSTAR's SLAM system, namely,
\emph{Large-scale Autonomous  Mapping  and  Positioning} (LAMP), is provided in~\Cref{fig:lamp_architecture}. \edited{LAMP is a key component of NeBula \cite{agha2021nebula}, team CoSTAR's overall autonomy solution.} LAMP relies on data from different odometry sources (i.e., \lidar, visual-inertial, wheel-inertial, and IMU) to estimate the robot trajectories, as well as a point cloud map of the environment. 
The system consists of (i) a \emph{single-robot front-end interface} that runs locally onboard each robot to produce an estimated robot trajectory and a point cloud map of the environment explored by each robot, (ii) a \emph{multi-robot front-end}, running on the base station, which receives the robots' local odometry and maps and performs multi-robot loop closure detection, and (iii) a \emph{multi-robot back-end}, that uses odometry (from all robots) and intra- and inter-robot loop closures from the multi-robot front-end to perform a joint pose graph optimization; the multi-robot back-end runs on the base station and simultaneously optimizes all the robot trajectories.  

\myParagraph{Single-Robot Front-End Interface} \label{sec:front-end-interface}
LAMP relies on a multi-sensor front-end interface that enables the use of robots with different sensor configurations and odometry sources, including LOCUS~\cite{andrzej2022iros} and Hovermap~\cite{jones2020applications}.
The front-end produces an odometric estimate of each robot's trajectory, and stores the corresponding information in a factor graph, where each node corresponds to an estimated pose, while an edge connecting two nodes encodes the relative motion between the corresponding timestamps. Each odometry node is associated with a keyed-scan, a pre-processed point cloud obtained at the corresponding timestamp. The keyed-scan are used for loop closure detection and to form a 3D map of the environment.

Within the single-robot front-end, LOCUS~\cite{andrzej2022iros} is CoSTAR's \lidar-centric odometry estimator.  
LOCUS starts with a pre-processing step, where ---after removing motion-induced distortions in point clouds---  scans from multiple onboard \lidars are merged into a unified point cloud given the extrinsic calibration between \lidars. An adaptive voxelization filter is then applied to ensure a constant number of points are retained independent of the environment geometry, point cloud density, and number of onboard \lidars. This helps  reduce the computation, memory usage, and communication bandwidth associated with the subsequent processing.
Odometric estimates are obtained using a two-stage scan-to-scan and scan-to-submap registration process; 
the registration relies on a fast implementation of point-to-plane ICP, initialized using IMU measurements or other odometry sources.

\myParagraph{Scalable Multi-robot Front-end}\label{sec:lidarLC}
The multi-robot front-end is in charge of intra- and inter-robot loop closure detection by leveraging a three-step process: loop closure generation, prioritization, and computation as outlined below.

The \emph{Loop Closure Generation} module relies on a modular design, where loop closure candidates can be identified using different methods and environment representations (i.e., Bag-of-visual-words~\cite{galvez2012bags}, junctions extracted from 2D occupancy grid maps \cite{ebadi2021dare}). 
 The go-to loop closure generation approach within \SubT has been based on \lidar point clouds.
 In particular, loop closure candidates are simply identified from nodes in the factor graph that lie within a certain Euclidean distance from the current node; the distance is dynamically adjusted to account for the odometry drift between nodes.  

The \emph{Loop Closure Prioritization} module~\cite{chris2022iros} selects the most promising loop closures for processing.
While loop closures are crucial for map merging and drift reduction in the estimated robot trajectory, it is equally crucial to avoid closing loops in ambiguous areas with high degree of geometric degeneracy \cite{ebadi2021dare}, as it could lead to spurious loop closure detections.
Furthermore, loop closure detection in large-scale environments, and with large number of robots, becomes increasingly more computationally expensive as the density of nodes in the pose graph, and subsequently the number of loop closure candidates, increases.
The purpose of this module is to prioritize loop closure candidates inserted in the computation queue by evaluating their likelihood of improving the trajectory estimate. This is achieved through a three-step process of (i) observability prioritization, where similar to the works presented in \cite{tagliabue2020lion, zhang2016degeneracy, ebadi2021dare}, eigenvalue analysis is performed to detect degenerate scan geometries, in order to prioritize loop closures in feature-rich areas, (ii) graph information prioritization, where a Graph Neural Network (GNN) \cite{zhou2020graph} based on a Gaussian Mixture model layer is used to predict the impact of a loop closure on pose graph optimization, and (iii) Receiver Signal Strength Indication (RSSI) prioritization to prioritize loop closures 
based on known locations indicated by RSSI beacons ---whenever a robot is within range of an RSSI beacon.
The prioritized loop closure candidates are inserted into a queue for the computation step in a round-robin fashion. 

The \emph{Loop Closure Computation} module estimates the relative pose between a pair of loop closure candidate nodes in the queue using a two-stage process.
First, an initial estimate of the relative pose is computed using TEASER++ \cite{yang2020teaser} or SAmple Consensus Initial Alignment (SAC-IA) \cite{rusu2009fast}. Then the Generalized Iterative Closest Point (GICP) algorithm \cite{segal2009generalized} is initialized with the obtained solution to refine the relative pose and evaluate the 
quality of the \lidar scan alignment.

\myParagraph{Robust Multi-robot Back-end}\label{sec:PGO}
LAMP uses a centralized multi-robot architecture, where a central base station receives the odometry measurements and keyed scans from each robot, along with loop closures from the multi-robot front-end, and performs pose graph optimization to obtain the optimized trajectory for the entire team. The optimized map is then generated by transforming the keyed scans to the global frame using the optimized trajectory.
To safeguard against erroneous loop closures, the multi-robot back-end includes two outlier rejection options: Incremental Consistency Maximization (ICM) \cite{ebadi2020lamp}, which checks detected loop closures for consistency with each other and the odometry before they are added to the pose graph, and Graduated Non-Convexity (GNC) \cite{yang2020graduated}, which is used in conjunction with the Levenberg-Marquardt solver to perform an outlier-robust pose graph optimization and obtain both the trajectory estimates and inlier/outlier decisions on the loop closure not discarded by ICM. 
Pose Graph Optimization and GNC are implemented using GTSAM~\cite{dellaert2012factor}.

\myParagraph{Dirty Details}
\textit{Parameter tuning:}
\edited{While LAMP provides} a robust localization and mapping framework, it is difficult to find a set of parameters for the front-end and back-end modules that leads to nominal performance consistently across environments with different topography and geometry. In order to have a more systematic approach to parameter tuning, CoSTAR curated 12 SLAM datasets across multiple challenging underground environments for evaluation and benchmarking, with the goal of obtaining at a set of parameters that gave the best performance across all domains. 
The parameter tuning was mostly manual, and was restricted to a small subset of parameters which had higher impact on the system's performance. 
One area where parameter tuning was successful was \lidar-based loop closure detection. Here, the dataset consisted of pairs of point clouds from a variety of environments, with 80\% of the pairs being true loop closures, with known relative poses, and the rest being outliers.

\subsection{Team CSIRO}
\label{sec:team-CSIRO}
\begin{figure*}[t]
		\centering
		\includegraphics[width=\textwidth]{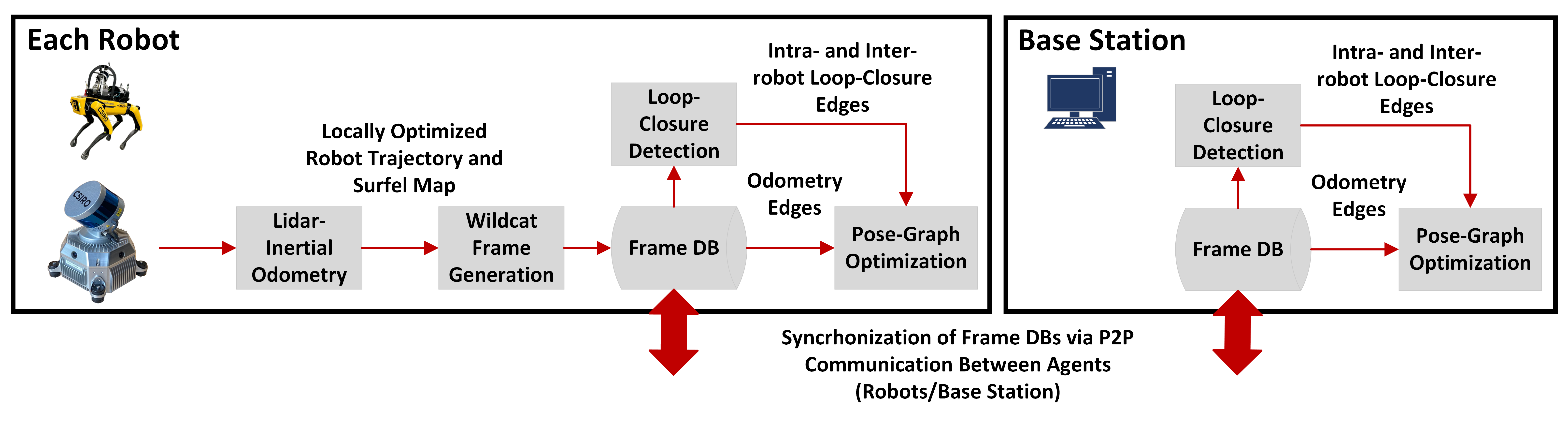}\vspace{-3mm}
		\caption{Overview of CSIRO's decentralized multi-robot SLAM system in \SubT. Each robot
		runs its own Wildcat's \lidar-inertial odometry module independently. The resulting
locally-optimized odometry estimate and surfel submaps are used to generate
Wildcat \emph{frames}. These frames are stored in a database and are shared with other
robots and the base station. Robots and the base station then use their
collection of frames to independently build and optimize a pose graph. \label{fig:CSIROflowchart}\vspace{-3mm}}
\end{figure*}

\edited{Team CSIRO Data61 tied for the top score and won the second place at the Final event of the DARPA \SubT challenge after the tiebreaker rules were invoked. 
The team also won the single most accurate artifact report award in the Urban and Final events.} 
An overview of \emph{Wildcat}~\cite{Wildcat,hudson2021heterogeneous}, CSIRO's \lidar-inertial decentralized multi-robot SLAM system, is given in~\Cref{fig:CSIROflowchart}. 
We first review CSIRO's distinctive sensing strategy, and then introduce the key modules in the Wildcat architecture: surfel generation, \lidar-inertial odometry, frame generation and sharing, and pose graph optimization.

\myParagraph{Sensing Pack} 
The ground robots carried a CatPack sensing payload designed by CSIRO. The CatPack uses an IMU and a Velodyne VLP-16 \lidar that is mounted at $45^\circ$ off horizontal and spins about the vertical axis of the CatPack. The CatPack also has four RGB cameras, which were used for artifact detection, but not for SLAM. The Emesent Hovermap~\cite{emesent} payload used on the aerial robots is a similar sensing pack with a spinning Velodyne VLP-16. Both the CatPack and Hovermap run the Wildcat SLAM system onboard, on their NVIDIA Jetson AGX Xavier and Intel NUC computers, respectively. 

The spinning \lidar configuration of CatPack provides dense depth measurements with an effective $120^\circ$ vertical field of view. This played a major role in making CSIRO's SLAM system robust in subterranean environments, \eg by providing improved visibility of the floor and roof of narrow tunnels. It also enabled use of surfel features, which exploit the dense depth measurements to provide a stable, robust feature set that is effective in a wide range of environments.

\myParagraph{Surfel Generation}
Wildcat uses planar surface elements (surfels) as dense features for estimating robot trajectory. Surfels are created every \SI{0.5}{s} by spatial and temporal clustering of new \lidar points. Specifically, the space is voxelized at multiple resolutions and points are clustered depending on their timestamp and the voxel they fall in. 
Clusters smaller than a predefined threshold (in terms of number of points) are discarded. An ellipsoid is then fit to each remaining cluster by computing the first two moments of its 3D points. The centroid (mean) of an ellipsoid specifies the position of the corresponding surfel, while its covariance matrix determines its shape. 
A planarity score \cite[Eq.\ 4]{bosse2009continuous} is computed based on the spectrum of the covariance, and only sufficiently planar surfels are kept.

\myParagraph{\Lidar-Inertial Odometry}
Wildcat's \lidar-inertial odometry module processes surfels and IMU data in a sliding window. \edited{Within a time window}, the processing alternates between (i) matching active surfel pairs and (ii) optimizing robot trajectory, for a predetermined number of times or until satisfying a convergence criterion. 
Surfel correspondences are established through $k$-nearest (reciprocal) neighbor search in the descriptor space comprising estimated surfel's position, normal vector, and voxel size. The estimate of the
segment of the
robot trajectory within the current time window is then updated by minimizing a cost function mainly composed of residual error functions associated to matched surfel pairs and IMU measurements in the current time window. The cost function is made robust to outliers (\eg incorrect surfel correspondences) by using the Cauchy M-estimator. 

\myParagraph{Frame Generation and Sharing}
A Wildcat \emph{frame} comprises a six-second portion of surfel map and odometry produced by each robot's \lidar-inertial odometry. Each robot generates {frames} periodically and stores them in a database. A frame is discarded if its surfel submap has very high overlap with that of the previous frame. As shown in~\Cref{fig:CSIROflowchart}, Wildcat leverages CSIRO's peer-to-peer ROS-based data sharing system, \emph{Mule}~\cite[Section 4.3]{hudson2021heterogeneous}, to synchronize robots' frame databases every time two agents (robot-robot or robot-base station) are within communication range.

\myParagraph{Pose Graph Optimization}
Each robot uses its collection of Wildcat frames \edited{---including those generated and shared by other robots---} to independently build and optimize the team's collective pose graph. Frames represent nodes of the pose graph. Each robot's odometry estimate is used to create odometry edges (i.e., relative pose measurements) between the robot's consecutive frames. Additionally, intra- and inter-robot loop-closure edges are created by aligning frames' surfel maps. This is done using ICP for nearby frames (for which a good initial guess is available from odometry) and global registration methods for distant ones. Pose graph nodes with significant overlap in their local maps are merged together. As a result, the computational complexity of the solver grows with the size of the explored environment rather than mission duration. The solver is made robust to outliers using Cauchy M-estimator. The collective pose graph built and optimized by each robot is used to render a surfel map of the environment.

\myParagraph{Dirty Details}
\textit{Parameter tuning:}
\edited{CSIRO's solution} uses a single set of parameters tuned to perform across a wide range of environments. 
However, ground robots and drones use different parameters due to the independent tuning processes.

\textit{Calibration:} CatPacks undergo extensive calibration on production, comprising both \lidar-IMU and \lidar-camera calibration. The incorporation of the cameras in the CatPack successfully avoided the need for subsequent calibration, even when packs are switched between platforms.  

\textit{Loop closures:} 
\edited{Complex \lidar-based place recognition techniques were rarely found to be necessary at the scale of \SubT environments, therefore team CSIRO found loop closures candidates by searching 
for past poses within a Mahalanobis distance from the current robot pose.}
In \SubT, the first inter-robot loop closures were created upon startup based on joint observation of the same starting region. This process at startup was imperfect, but difficulties could be addressed procedurally, \eg by restarting the affected agent. Since each agent also solves independently for its own multi-robot solution, it was necessary to ensure that these inter-robot loop closures are successfully detected not only on the base, but also on each robot. After difficulties in the Urban Event of the competition, user interface elements were introduced to prominently report the connectivity status of the collective pose graph to detect anomalies. By the Final event, these failures were rare.

\textit{Hardware robustness:} While hardware robustness is \edited{seldom} discussed in the SLAM literature, it is a significant feature of the CSIRO system's maturity. On the rare occasions when Wildcat diverged in testing, almost all occurrences were found to coincide with sensor dropouts caused by significant kinetic impacts of the platform, or hardware failures, which typically start with intermittent errors.

\subsection{Team CTU-CRAS-Norlab}
\label{sec:team-ccn}


\begin{figure*}[t!]
    \centering
    \includegraphics[trim={0cm 0cm 0cm 0cm},clip, width=0.9\textwidth]{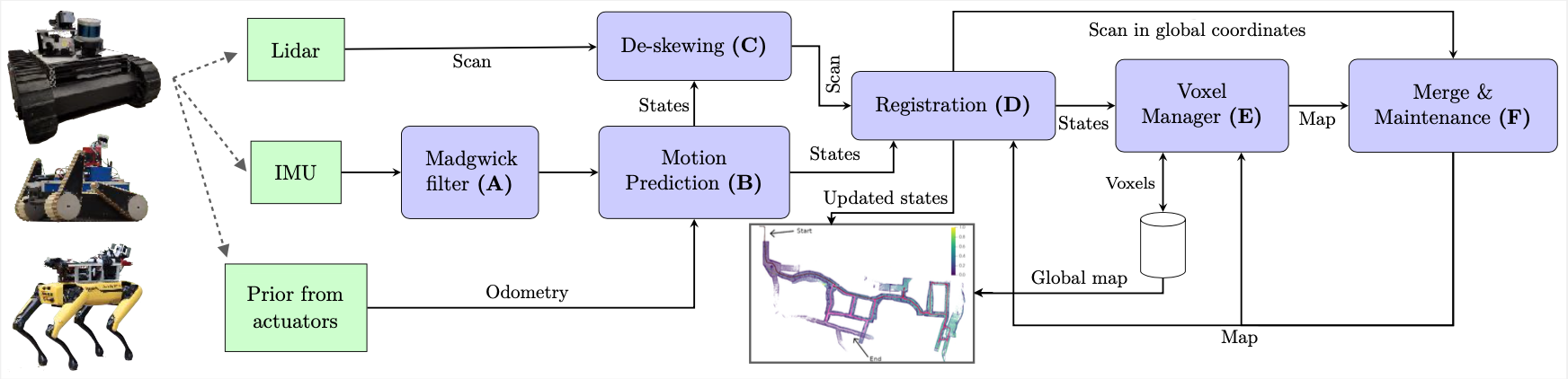}
    \caption{CTU-CRAS-Norlab UGV SLAM architecture (\edited{Norlab ICP Mapper}). \edited{Green boxes correspond to the inputs \LC{and output}, and the purple ones represent submodules of the SLAM architecture.}\label{fig:UGV_architecture} \vspace{-4mm}} 
\end{figure*}

\begin{figure}[tb]
  \centering
  \begin{tikzpicture}[node distance=2cm]

    \pgfmathsetmacro{\vshift}{0.07em} 
    \pgfmathsetmacro{\hshift}{0.14em} 
    \pgfmathsetmacro{\vblocksize}{1.9} 
    \pgfmathsetmacro{\hblocksize}{2.0} 
    \pgfmathsetmacro{\harrowshiftem}{0.8em} 
    \pgfmathsetmacro{\varrowshiftem}{0.6em} 
    \pgfmathsetmacro{\varrowshift}{0.01em} 
    \pgfmathsetmacro{\harrowshift}{0.03em} 
    \pgfmathsetmacro{\harrowbend}{0.38mm} 
    \pgfmathsetmacro{\varrowbend}{0.03em} 


    \node[block, shift = {(0.0, 0.0)}] (buffer) {
      \begin{tabular}{c}
        \small History buffer
      \end{tabular}
    };

    \node[block, below of=buffer, shift = {(0.0, 1.5*\vshift)}] (kalman_filters) {
      \begin{tabular}{c}
        \small Kalman filters
      \end{tabular}
    };

    \node[block, left of=buffer, shift = {(-0.4*\hshift, 0.0)}] (aloam) {
      \begin{tabular}{c}
        \small LOAM\\
      \end{tabular}
    };

    \node[block, below of=aloam, shift = {(0.0, 1.5*\vshift)}] (mapping) {
      \begin{tabular}{c}
        \small Mapping\\
      \end{tabular}
    };

    \node[block, above of=aloam, shift = {(-0.0, -1.5*\vshift)}] (3d_lidar) {
      \begin{tabular}{c}
        \small 3D Lidar\\
      \end{tabular}
    };

    \node[block, left of=3d_lidar, shift = {(-0.0*\hshift, 0.0)}] (filtration) {
      \begin{tabular}{c}
        \small Filtration\\
      \end{tabular}
    };

    \node[block, left of=aloam, shift = {(0.0, -0.0*\vshift)}] (depth) {
      \begin{tabular}{c}
        \small Depth cam\\
      \end{tabular}
    };

    \node[block, left of=mapping, shift = {(-0.0*\hshift, 0.0)}] (filtration_depth) {
      \begin{tabular}{c}
        \small Filtration\\
      \end{tabular}
    };
    
    \node[block, right of=buffer, shift = {(0.2*\hshift, 0.0*\vshift)}] (imu) {
      \begin{tabular}{c}
        \small IMU\\
      \end{tabular}
    };

    \node[below of = kalman_filters, shift = {(1.5*\hshift, 2.0*\vshift)}] (output) {
    };



    \draw [arrow, color=black] (3d_lidar.west) -- (filtration.east) node[midway, above] {\small $\mathcal{P_L}$} ;
    \draw [arrow, color=black] (filtration) |- ($(filtration.south) + (0, -0.30)$) -| ($(aloam.north) + (0, 0.20)$) -- (aloam) node[midway, above, shift={(-0.7*\hshift, 0.05*\vshift)}] {\small $\mathcal{P}_{LF}$ } ;
    \draw [arrow, color=black] (filtration) |- ($(filtration.south) + (0, -0.30)$) -| ($(filtration.east) + (0.7*\harrowshift, -0.55)$) |- (mapping.172) ;
    \draw [arrow, color=black] (aloam) -- (mapping) node[midway, left] {\small $\mathbf{r}$} ;
    \draw [arrow, color=black] ($(aloam.east) - (0, 0)$) -- ($(buffer.west)$) node[midway, above, shift={(-0.0, 0)}] {\small $\mathbf{r}$ };
   \draw [arrow, color=black] (imu.west) -- node[midway, above, shift={(0.05, 0.02)}] {\small $\mathbf{\dot{R}}, \mathbf{\ddot{r}}$ } (buffer.east) ;

    \draw [darrow, color=black] ($(buffer.200)$) -- ($(kalman_filters.160)$) node[midway, left] {\small $\mathbf{s}$} ;
    \draw [arrow, color=black] ($(buffer.340)$) -- ($(kalman_filters.20)$) node[midway, left] {\small $\mathbf{z}$} ;

    \draw [arrow, color=black] (kalman_filters.south) |- (output.west) node[midway, above, shift={(1.5, 0)}] {\small $\mathbf{s}$};
    \draw [arrow, color=black] (mapping.south) |- ($(output.west) + (0, -0.2)$) node[midway, above, shift={(1.5, 0)}] {\small $\mathcal{M}$};
    \draw [arrow, color=black] (depth) -- (filtration_depth) node[midway, left] {\small $\mathcal{P_D}$} ;
    \draw [arrow, color=black] ($(filtration_depth.-7) - (0, 0)$) -- ($(mapping.187)$) node[midway, below, shift={(-0.0, 0)}] {\small $\mathcal{P_{DF}}$ };

  \begin{pgfonlayer}{background}
    \path (buffer.west |- buffer.north)+(-0.10,0.4) node (a) {};
    \path (kalman_filters.south -| kalman_filters.east)+(+0.05,-0.1) node (b) {};
    \path[rounded corners, draw=black!70, densely dotted]
    (a) rectangle (b);
  \end{pgfonlayer}
  \node [rectangle, above of=buffer, node distance=1.9em, shift={(0.00,-0.20)}] (node_text) {\footnotesize \textbf{State estimation}};

  \end{tikzpicture}
  \caption{CTU-CRAS-Norlab UAV SLAM architecture.
  $\mathcal{P_D}$ and $\mathcal{P_L}$ are the depth camera and 3D \lidar point clouds. $\mathcal{P_{DF}}$ and $\mathcal{P_{LF}}$ are the respective point clouds after filtration.
  The outputs are the map $\mathcal{M}$ and the state $\mathbf{s}$, which consists of position $\mathbf{r}$, orientation $\mathbf{R}$, and their first derivatives $\mathbf{\dot{r}}$ and $\mathbf{\dot{R}}$. 
  The Kalman filter corrections $\mathbf{z}$ consist of $\mathbf{r}$, $\mathbf{\dot{R}}$, and $\mathbf{\ddot{r}}$.
  \label{fig:UAV_architecture} \vspace{-5mm}}
\end{figure}
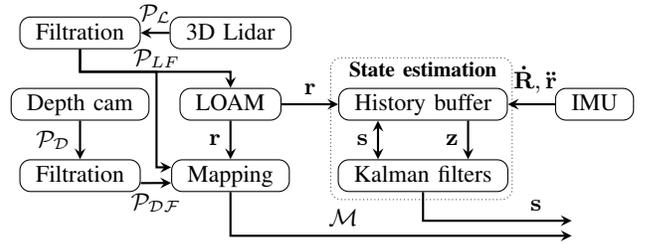

The CTU-CRAS-Norlab team employed two separate {SLAM} systems for their \ac{UGVs} and \ac{UAVs}. 
The corresponding architectures are given in~\Cref{fig:UGV_architecture} and~\Cref{fig:UAV_architecture}, respectively.

\myParagraph{UGV SLAM}
The {UGV SLAM} architecture relies exclusively on \edited{a \lidar odometry system}, \edited{Norlab ICP Mapper}, that focuses on reducing drift at the front-end level.
The \edited{mapper} operates as follows: (A) first, the robot orientations during a \lidar scan are estimated by passing the {IMU} measurements through a Madgwick filter~\cite{madgwick2011estimation}.
(B) then, this orientation information is fused with translation estimates from wheel odometry to estimate the robot motion during the scan.
(C) the motion estimate allows to de-skew the current \lidar scan (\ie motion correction).
(D) once de-skewed, the \lidar scan is registered in the local map using {ICP}, taking the robot pose as prior.
A modified version of point-to-plane {ICP}\edited{~\cite{Kubelka2022}} is used, where only 4 degrees of freedom (3D position and yaw angle of the scan) are optimized, \edited{while roll and pitch angles are directly obtained from the IMU.}
(E) the robot pose found using registration, is used by the voxel manager to load and unload voxels of the local map to ensure it stays centered on the robot.
(F) lastly, the registered cloud is merged into the local map and maintenance operations are performed.
These maintenance operations include identifying and removing points belonging to dynamic objects using the technique described in~\cite{pomerleau2014long}.
The resulting map is then set as the new local map.
These steps are performed in different threads to allow the system to localize at a higher rate than the rate at which the map is updated. 

\myParagraph{UAV SLAM}
The UAV SLAM architecture relies on a \lidar sensor that is complemented by an IMU for precise roll-and-pitch orientation estimation. 
While not necessary for localization, which utilizes only \lidar and IMU measurements, data from upward- and downward-facing depth cameras are integrated into a dense metric map to cover the blind spots of the \lidar field of view.
The output of the system is a state estimate (\ie robot poses in a gravity-aligned reference frame, and their derivatives), and a volumetric occupancy map.

The UAV SLAM pipeline (\Cref{fig:UAV_architecture}) starts with pre-processing of \lidar scans.
First, a range-clip filter is applied to the raw scans to filter out the robot frame and distant measurements.
Second, a local intensity-threshold filter is applied to the data, which proved to be a highly robust method for filtration of dust even in the harshest conditions.
Due to computational constraints, this pre-processing does not involve \lidar scan de-skewing; while this negatively impacts the SLAM performance, it reduces the delay incurred by the pose estimate.
The processed data is passed to LOAM~\cite{zhang2014loam}, which optimizes the alignment of geometric features extracted from the data in a two-step odometry process --- fast scan-to-scan and slow scan-to-map matching in the feature space.
The team has adapted the advanced implementation of LOAM (A-LOAM\footnote{\scriptsize{\url{https://github.com/HKUST-Aerial-Robotics/A-LOAM}}}) to be suitable for UAVs 
by extending the method with platform-optimized parallelization.
The state estimation module (based on~\cite{petrlik2020robust}) takes the LOAM pose estimate 
and fuses it with the IMU measurements using a linear Kalman filter to obtain a high-rate delay-compensated state estimate that is suitable for the control system feedback loop~\cite{baca2021mrs}.
The non-constant delay introduced by LOAM negatively impacts the controller performance and, most importantly, the control error.
The idea of the delay-compensation method~\cite{pritzl2022repredictor} is to recompute the current state if a measurement with a past timestamp arrives.
When a delayed measurement arrives, it is applied to the state in a circular buffer with the nearest timestamp.
The corrected state is then propagated to the current time using the system model and other relevant updates.

The approach is fully decentralized, with each UAV running its own SLAM pipeline.
To allow multi-robot cooperation, the reference frames of UAVs are initially aligned by one of the following procedures.
The reference frames are either given in advance (\eg from a total station) or their alignment is estimated with respect to a leader-robot by scan matching using \lidar data shared among all the UAVs before takeoff.

\myParagraph{Dirty Details}
\textit{Loop closures:} 
The only tunable parameter of the UAV SLAM method is the resolution of the feature map. For both SLAM systems, it was found empirically that one set of parameters worked well in a majority of scenarios; any changes to the parameters led to degraded state estimation quality or slower-than-real-time performance. Not adapting the parameters dynamically also ensured static assignment of computational resources, which helped to predict and optimize the system behavior under resource constraints.

\textit{Loop closures:} 
Neither the UGV nor the UAV SLAM systems detect loop closures; therefore, no pose graph optimization is used to refine the odometric trajectories.

\textit{Computation prioritization:} 
When the CPU is fully loaded, critical components \edited{such as control and SLAM} might have to wait for CPU resources shared with non-flight-critical software \edited{such as object detectors}, which results in triggering failsafe recovery behaviors. Prioritizing the critical modules at the process level by reducing CPU affinity and using negative \verb|nice|\edited{~\cite{nice}} values for non-critical processes resulted in lower computation times, lower jitter, and smoother flights. Additional performance was gained by running the algorithms that process large amounts of data as nodelets under a common ROS nodelet manager. This avoids the overhead of copying large data structures by simply passing pointers instead.

\subsection{Team Explorer}
\label{sec:team-explorer}

\begin{figure*}[t!]
    \centering
    \includegraphics[trim={0cm 0cm 0cm 0cm},clip, width=0.7\textwidth]{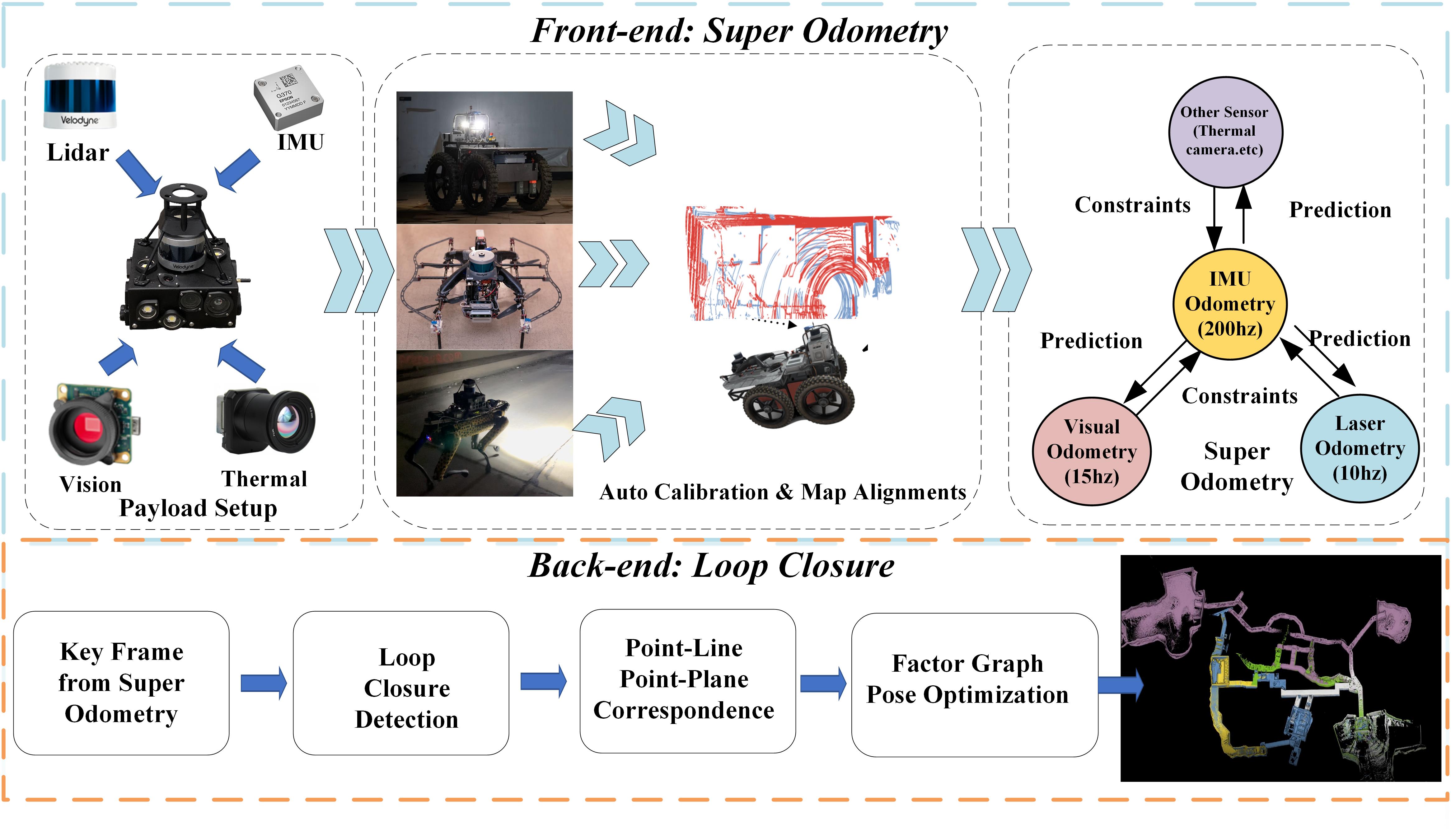}\vspace{-3mm}
    \caption{Overview of team Explorer's SLAM architecture. Each robot estimates and operates on its own individual map.\label{fig:explorer_architecture} \vspace{-3mm}}
\end{figure*}

Team Explorer won the Tunnel event of the DARPA \SubT Challenge.
Team Explorer's SLAM architecture is given in~\Cref{fig:explorer_architecture}. 
The architecture relies on Super Odometry~\cite{zhao2021super} to fuse outputs of multiple odometry sources including visual or thermal fusion \cite{TP-TIO} using a probabilistic factor graph optimization, and a loop-closing back-end.

\myParagraph{Lidar-Inertial Localization Module for Odometry Estimation}
The lidar-inertial localization module relies on Super Odometry (SO)~\cite{zhao2021super}. 
In SO, a factor graph optimization performs estimation over a sliding window of recent states by combining IMU pre-integration factors with 
point-to-point, point-to-line, and point-to-plane \lidar factors.
SO strikes a balance between loosely- and tightly-coupled estimation methods.
The IMU-centric sensor fusion architecture  does not combine all sensor data into a full-blown factor graph. Instead, it breaks it down into several ``sub-factor-graphs'', with each one receiving the prediction from an IMU pre-integration factor.  The motion from each odometry factor is recovered in a coarse-to-fine manner and in parallel, which significantly improves real-time performance. SO enables achieving high accuracy and operates  with a  low  failure  rate, since the IMU sensor is environment-independent, and the architecture is highly redundant. As long as  other  sensors  can  provide  relative  pose  information  to constrain  the  IMU  pre-integration  results,  these  sensors will be fused into the system successfully. 

\myParagraph{Loop-Closing Back-End}
While SO is a low-drift odometry algorithm, it is still important for the SLAM system to be able to correct long-term drift. This is specially true when the traversed distance is high. Considering that Explorer's ground robots moved at
\SI{0.5}{\metre\per\second} and had an aggressive exploration style, it was not uncommon to observe traversed distances larger than \SI{1}{\km} in a single test.
Team Explorer's solution reduces the drift by detecting loop closures and performing pose graph optimization.
In particular, the back-end filters the poses and point clouds generated by the front-end. It applies a heuristic method to accumulate these results into a \textit{keyframe}, which is composed by a \textit{key pose} and a \textit{key cloud}, which is the point cloud generated by accumulating all the point clouds generated since the last keyframe, and downsampling to maintain a fixed size. The heuristic used is distance-based: a new keyframe is created after the robot moves by \SI{0.2}{\metre}. Search for loop closures is performed using a radius-search among the nearest poses, or by querying a database of sensor data to find matches with previously visited places.

\myParagraph{Autocalibration}
To achieve a common task, multiple robots need to be able to establish and operate in a common frame of reference. 
Towards this goal, team Explorer used \textit{Autocalibration}, a process in which a Total Station is used to obtain the pose of one robot with respect to the fiducial markers \edited{with known positions in the world frame} set by DARPA. 
This robot shares its pose with respect to its own map frame and the latest three keyframes it created. All the following robots will then be placed near the calibration location of the first robot; they receive the reference information from the base station, and use GICP \cite{segal2009generalized} to align their current keyframes to establish their initial pose in the world frame.

\myParagraph{Dirty Details}
\textit{Loop closures:} 
\edited{The most important parameters tuned} during testing were those related to the downsampling of the point clouds before the scan-to-map registration. This downsampling affected the number and the quality of features available to SO. 
In particular, a key parameter is the voxel size used in the PCL library voxel grid filter.
When the robot traverses a narrow urban corridor, it is desirable to use a smaller voxel size, to avoid decimating important details in the point cloud. In contrast, in a large cave, a larger voxel size is required, otherwise the processing becomes too slow due to the large number of features.
To solve this problem, we created an heuristic method to switch voxel sizes in real-time.
The method consists in calculating, for each 3D axis separately, the average distance to the points in the current point cloud. Then, we multiplied the 3 values together to obtain an ``average volume''. This volume was thresholded to create 3 different modes, each associated to a predefined voxel size. 

\textit{Dust Filters:}
Team Explorer had a strong focus on UAVs. These platforms bring their own unique challenges to the SLAM problem. One that was particularly important for \SubT was being able to handle the dust that arises due to the robot's propellers. The simple solution that was implemented was to test if there was a minimum number of features farther than \SI{3}{\metre} from the robot. If true, all the other points inside this radius were ignored when performing pose estimation. This solution was based on the assumption that dust would usually accumulate circularly around the robot, but usually there are still other distant features in the environment that allow the robot to solve the optimization correctly. 
If the robot were capable of performing estimation and continuing operation, it would usually escape the dusty area 
in the environment. Otherwise, dust would eventually cover the robot from all sides and a catastrophic failure would occur.

\textit{Robot-specific computational budget:}
Analysis of empirical results showed that after a certain number, having more \lidar features does not necessarily translate to substantial accuracy gains. Therefore, a threshold on the number of surface features is used. If the current scan frame contains more than the threshold, the list of features is sampled uniformly such that the number of features does not exceed the threshold.


\subsection{Team MARBLE}
\label{sec:team-MARBLE}

\begin{figure*}[t!]
 \centering
 \includegraphics[width=0.8\linewidth]{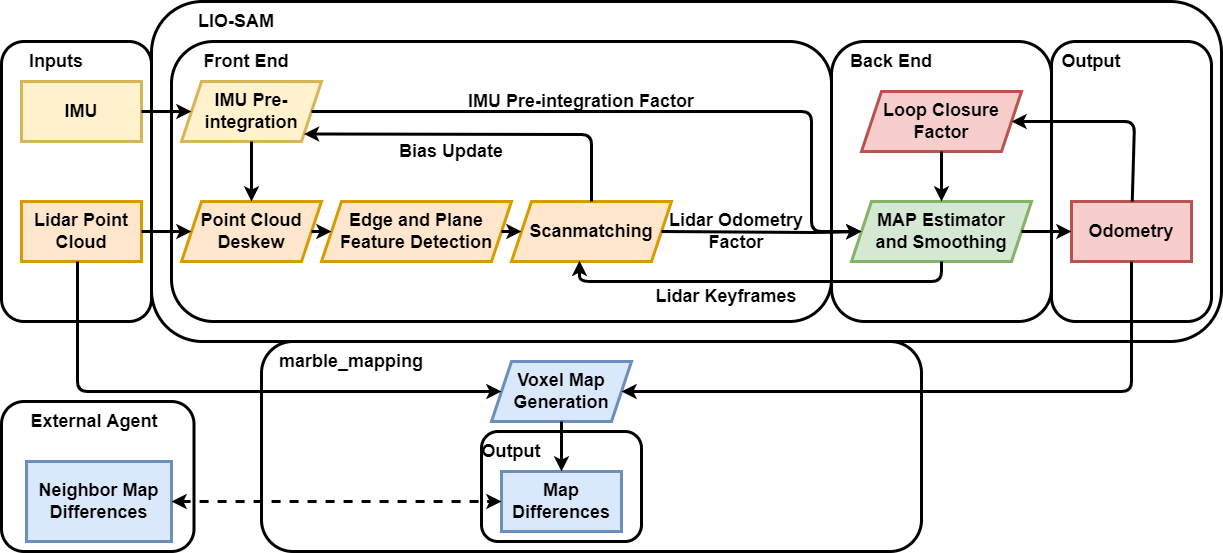}
 \caption{Overview of team MARBLE's SLAM architecture. \label{fig:marble_architecture} \vspace{-5mm}}
\end{figure*}

Team MARBLE's SLAM architecture is given in~\Cref{fig:marble_architecture}.
The core of MARBLE's \lidar-centric solution\footnote{Initially, MARBLE explored the use of onboard cameras and a visual-inertial odometry system~\cite{kramer2021vi}, 
namely, Compass~\cite{nobre2016multi}. While this solution was tenable in some cases, changing lighting conditions and specular highlights caused by on board illumination in dark scenes often lead to instability. 
This was especially true in longer deployments, such as the hour-long runs necessitated by \SubT. A dataset for benchmarking visual-inertial SLAM systems with onboard illumination was released as a part of these tests \cite{kasper2019benchmark}, however team MARBLE switched to a \lidar based architecture soon after.}  is the open-source LIO-SAM~\cite{shan2020lio} package, which performs tightly-coupled fusion of IMU data and LOAM-based  \lidar features~\cite{zhang2014loam}. The localization results are then passed to \edited{MARBLE Mapping}, {that creates a voxel map.

\myParagraph{\lidar Localization via LIO-SAM}
Each robot in the MARBLE system was responsible for its own localization, from input (\ie \lidar scans \edited{at \SI{20}{Hz}} and IMU data at \SI{500}{Hz}) to optimization. 
The localization process includes multiple subcomponents (\Cref{fig:marble_architecture}). First of all, in order to be processed by LIO-SAM,
each point in the \lidar point cloud required two extra data fields in addition to the standard $x,y,z$ position: 
a timestamp, and a ring number to provide their relative position in the vertical scan. 
This additional data is used to de-skew the point clouds.
\edited{While current Velodyne LIDAR and Ouster-OS1 LIDAR drivers provide this information by default, the Ouster-OS1 \lidar driver required some slight modification to enable this information.} In particular, \edited{timestamps} were added to each vertical angle of arrival, and rings were designated by their elevation angle. 

Localization via LIO-SAM is based on factor graph optimization and involves three types of factor. 
The first type consists in IMU pre-integration factors~\cite{Forster17tro}. The second type includes \lidar odometry factors; in particular, 
once the \lidar has been de-skewed, LIO-SAM extracts key features along lines and edges (as in LOAM~\cite{zhang2014loam}). These features are then compared and scan-matched along a subset of local key frames in a sliding window filter. Lastly, loop closure factors are determined by a naive Euclidean distance metric. 
Each time a new factor is added to the graph, the iSAM2 solver~\cite{kaess2012isam2} is applied to optimize the graph using GTSAM~\cite{dellaert2012factor}. 
\edited{After generating an odometry estimate, LIO-SAM then estimates the IMU bias with the updated odometry.}

{\bf Multi-Robot Mapping via Octomap.}
LIO-SAM outputs robot pose estimates. A voxel based map can also be queried via a ROS service call, however, MARBLE relied on a separate custom package, \edited{MARBLE Mapping}, a fork of Octomap \cite{hornung2013octomap}, which allowed creating voxel grid map differences with low data transfer requirements. 
In particular, \edited{MARBLE Mapping} uses the latest LIO-SAM pose estimate and 
the corresponding \lidar scan to update the log-odds probability (occupancy) value inside an octomap  \edited{with voxel size of \SI{0.15}{\metre}}.}
When enough voxels have been added or have changed state, or if enough time or distance has been traversed, a new map difference is created by the robot with the changed voxels. 

Map differences are then shared between robots in a peer-to-peer fusion. Each robot tracks differences in a sequence tied to its own \edited{identifier} and \edited{that} of their neighbors. Then when two robots connect to each other or the base station via a deployed mesh network, they request any differences not contained with their own \edited{maps}, and pass on any differences they had generated to the neighboring agent. To minimize overhead, maps were transmitted in their binary state, 
after thresholding the occupancy probability to an occupied/unoccupied state.

As each robot only optimized its own trajectory and map, any significant drift or misalignment between robots could cause potential downstream issues with multi-agent planning algorithms. To mitigate this issue, each robot prioritizes its own map, specifically during the merging process, where free voxels in the parent robot were kept free and the occupied voxels were merged together. The base station operator also had the ability to remove or stop merging differences from specific agents if significant tracking errors occurred.

\myParagraph{Dirty Details}
\textit{Parameter tuning and IMU:}
 \edited{The IMU covariance} was found to have a substantial impact on the roll, pitch, and yaw estimation. In constrained passage ways, rotation accuracy significantly decreased as a result of a significant number of \lidar points falling below a minimum range threshold. Relying more heavily on the IMU during these maneuvers improved rotation accuracy substantially (although it did not fully eliminate the problem). In this regard, using a good IMU is paramount: the LORD Microstrain 3DM-GX5-15, provides exceptionally high accuracy pitch and roll estimates of \SI{0.4}{\degree}, along with a \SI{0.3}{\degree\per\sqrt{\text{hr}}} gyro estimate~\cite{lord_data_sheet}, which allowed the MARBLE system to rely on IMU-only measurements for extended periods of time.

A second key parameter in the system was the key-frame search radius for loop closures. Given the localization maintained qualitatively good accuracy, the Euclidean search distance was continually reduced, resulting in a final distance of \SI{2}{\metre} for loop closure constraints. As loop closure optimization were computationally expensive, this saved on CPU cycles and additionally helped avoid spurious loop closure between different elevations of tunnels or floors in a building. 

\textit{Hardware design:}
Team MARBLE relied on precision machining to obtain (and preserve) an accurate extrinsic calibration between \lidar and IMU. Further calibration may have benefited the final solution ---specifically, improved IMU noise and bias estimation. It was found that certain IMUs did not perform as well as others in qualitative analysis of two robots traversing roughly the same trajectories. The team opted to swap hardware over further exploration of the cause of these errors. The chosen hardware likely had the closest noise parameters to those provided by the IMU manufacturer. 

\textit{LIO-SAM enhancements:}
Team MARBLE also made two minor adjustments to LIO-SAM. During initialization, the team chose to ignore measurements from the IMU until a point cloud had been received, since the IMU was not a full AHRS unit and did not have a heading compass. 
The second adjustment was to the IMU \edited{timestamp}s, prior to integration. As a result of the (ACM-based) USB driver used by the IMU, the  measurements did not have guaranteed priority on the kernel. This meant the \edited{timestamp}s generated by the system were not always consistent, which often caused negative \edited{timestamp} values in the IMU pre-integration, leading to instabilities. 
To avoid this issue, the MARBLE implementation replaced the \edited{timestamp}s (using the nominal IMU frequency) if they were outside an acceptable range. While this method was less precise than a full hardware clock sync, it was fairly easy to implement given the available onboard connections (a full hardware sync would have required an extra RS232 port).
In practice, we noticed that the back-end optimizer was able to mitigate the impact of minor \edited{timestamp} mismatches.


\subsection{Common Themes on the Path to Robustness}
\label{sec:common-themes}

\begin{table*}[b!]
  \centering
  \caption{Open-source Datasets and Code Released by the \SubT teams}
  \label{tab:code_and_dataset}
  \begin{tabular}{|c|c|c|}
  \hline
  \textbf{Team}       &\textbf{ Code}  & \textbf{Dataset}  \\ \hline
  \textbf{CERBERUS}     &
  \multicolumn{2}{c|}{\url{https://www.subt-cerberus.org/code--data.html}} \\
  \hline
  \textbf{CoSTAR}      &
  \multicolumn{2}{c|}{\url{https://github.com/NeBula-Autonomy}} \\
  \hline
  \textbf{CTU-CRAS-Norlab}  & \url{https://github.com/ctu-mrs/aloam} & \\ &
  \url{https://github.com/ctu-mrs/octomap_mapping_planning} & \\ &
  \url{https://github.com/ctu-mrs/mrs_uav_system} & \\ &
  \url{https://github.com/norlab-ulaval/norlab_icp_mapper} & \url{https://github.com/ctu-mrs/slam_datasets} \\ & 
  \url{https://github.com/norlab-ulaval/norlab_icp_mapper_ros} & \\ & \url{https://github.com/ethz-asl/libpointmatcher} & \\ & \url{https://github.com/norlab-ulaval/libpointmatcher_ros} & \\ & \url{https://github.com/ethz-asl/libnabo}  & \\ 
  \hline
  \textbf{Explorer}   &\url{https://www.superodometry.com/}& \url{https://theairlab.org/dataset/interestingness} \\
   &\url{https://theairlab.org/research/2022/05/02/subt_code/} &\url{https://www.superodometry.com/datasets} \\\hline
  \textbf{MARBLE}      & \url{https://github.com/arpg/marble_mapping} & \url{https://arpg.github.io/coloradar} \\ &
  \url{https://github.com/arpg/LIO-SAM} & 
  \url{https://arpg.github.io/oivio} \\ \hline
  \end{tabular}
\end{table*}

Despite the unique features that distinguish the architectures adopted by the \SubT teams, 
the previous sections reveal a substantial convergence of technical approaches across teams. 
This convergence is a testament of the maturity of multi-robot \lidar-centric SLAM, \edited{at least for small robot teams, (\eg 5-10 robots)}. We discuss commonalities across systems below.

\myParagraph{Sensing}
\edited{Most} teams relied on \lidar and IMU as the dominant sensing modalities; 
IMUs are not sensitive to perceptual aliasing (\ie the case where different places have the same appearance/sensor footprint) and environmental disturbances; \lidars afford accurate and long-range depth measurements even in the absence of external illumination. 
At the same time, visual, thermal, and wheel/kinematic odometry remain an important addition to \lidar, especially in the presence of obscurants and to increase redundancy. 
Many teams (\eg CSIRO, Explorer, MARBLE, CoSTAR)  adopted a common sensor payload to be mounted on the different robots. This modular design allows standardizing calibration and testing procedures, and partially decouples the development of the SLAM system from other hardware choices.

\myParagraph{SLAM Front-end and Back-end}
\edited{All} teams relied on local (single-robot) front-ends to pre-process the \lidar data. 
Such pre-processing reduces the data volume communicated to the base station or to the other robots.
Moreover, it allows splitting computation across the robots, improving scalability. 
Most solutions perform extensive point cloud pre-processing, including de-skewing and voxel grid filtering.
The front-ends then process the \lidar scans using feature-based (akin to LOAM~\cite{zhang2014loam}) or dense (\eg ICP-based) matching.
Regarding the SLAM back-end, virtually all teams relied on factor graph or pose graph optimization 
(\edited{except for} the Kalman-filter-based odometry from CTU-CRAS-Norlab).
Several teams decided not to detect loop closures (\eg CTU-CRAS-Norlab, and partially CERBERUS), based on considerations about the scale of the environment and the computational constraints at the robots. 
Finally, most teams built on top of open-source libraries for the \lidar front-end and back-end, 
including GTSAM~\cite{dellaert2012factor}, maplab~\cite{schneider2018maplab}, LOAM~\cite{zhang2014loam}, LIO-SAM~\cite{shan2020lio}, Octomap~\cite{hornung2013octomap}, \edited{and libpointmatcher~\cite{Pomerleau2013b}}.

\myParagraph{Loosely-coupled vs. Tightly-coupled Architectures} 
 Most teams resorted to loosely-coupled sensor fusion techniques, where estimates from multiple sensors are first fused into pose estimates and then combined together. 
Loosely-coupled approaches enable a more modular software design and make the implementation of health checks for each data source and intermediate pose estimate easier. This has been shown to largely increase robustness to hardware and software failures, \eg~\cite{zhao2021super,palieri2020locus,khattak2020complementary}.
In addition, tightly-coupled fusion leads to larger optimization problems, which prevents scaling the multi-robot back-ends to large teams.

\myParagraph{Centralized and \edited{Decentralized Architectures}}
CERBERUS and CoSTAR adopted centralized architectures, where the base station performs a joint optimization over the entire robot team.
All the other teams adopted a \edited{decentralized} approach, where each robot mostly operated on its own, with the occasional exchange of the mapping results (see CTU-CRAS-Norlab, Explorer, and MARBLE) 
or with a multi-robot pose graph optimization executed at each robot (CSIRO). 
No team adopted a distributed architecture, which are still the subject of active research~\cite{Tian21tro-KimeraMulti} and were less amenable to the rules of the \SubT competition, which required collecting data at a base station for visualization and scoring purposes.  
\omitted{
\myParagraph{Unconventional Features}
Despite the similarities, each team's solution is unique in some aspect. 
Among many other unique features discussed in the previous sections, we recall 
(i) the intensity-based dust filter, thread prioritization, and 4-dof ICP used by CTU-CRAS-Norlab; 
(ii) CSIRO's spinning \lidar solution (CatPack);
(ii) the loop prioritization and outlier rejection by CoSTAR;
(iv) the adaptive parameter tuning (auto-resolution) approach by team Explorer;
(v) MARBLE
CERBERUS
}


\begin{figure}[b!]
\centering
	\includegraphics[width=1.0\columnwidth, trim= 0mm 0mm 0mm 1mm, clip]{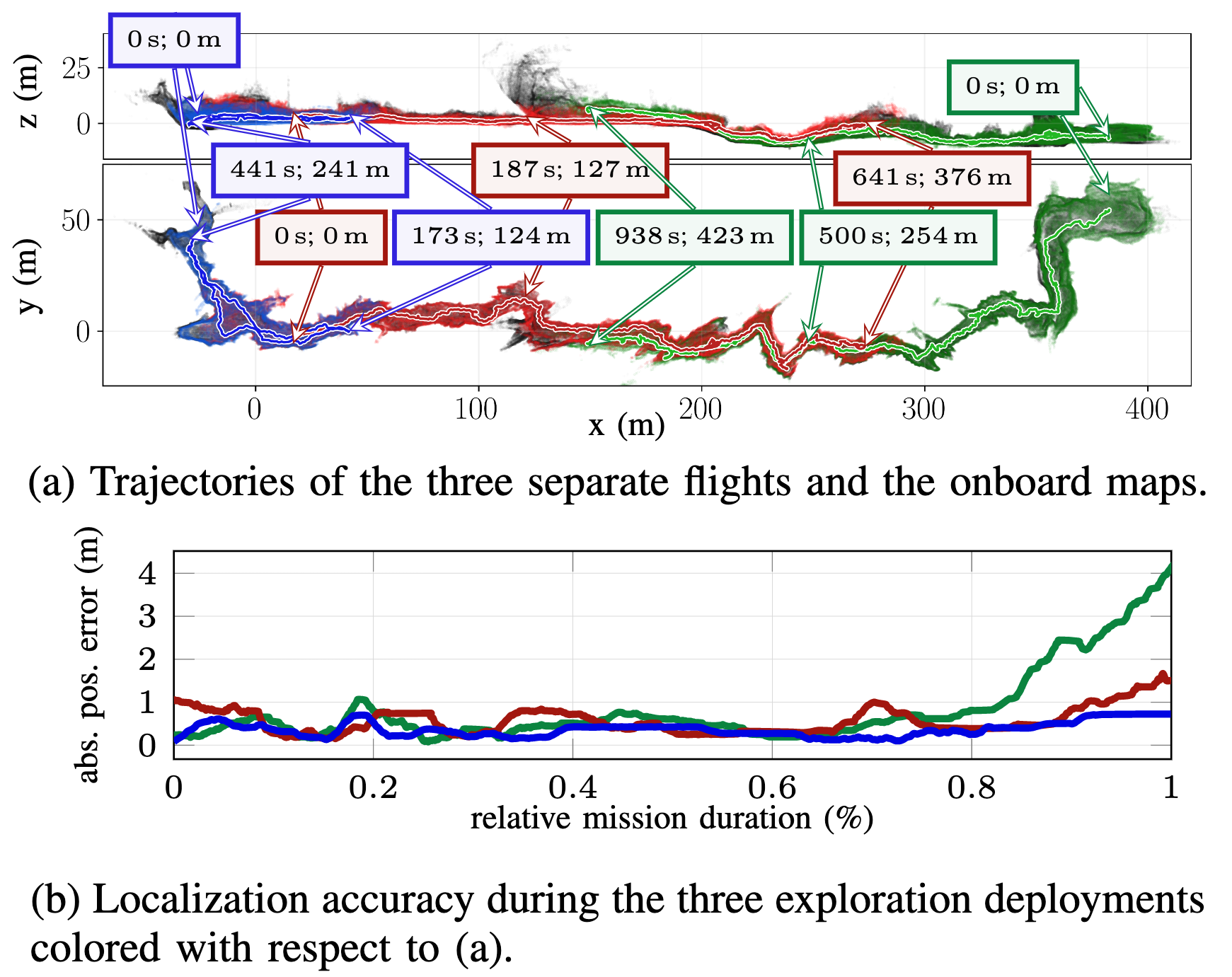}
	\caption{CTU-CRAS-Norlab's odometry accuracy for three single-UAV deployments in the Bull Rock Cave system~\cite{petracek2021caves}
	\label{fig:CTU_CRAS_NORLAB_UAV_Localization_Error}\vspace{-3mm}
	}
\end{figure}

\begin{figure}[t!] 
\centering
	\includegraphics[width=1.0\columnwidth, trim= 0mm 0mm 0mm 1mm, clip]{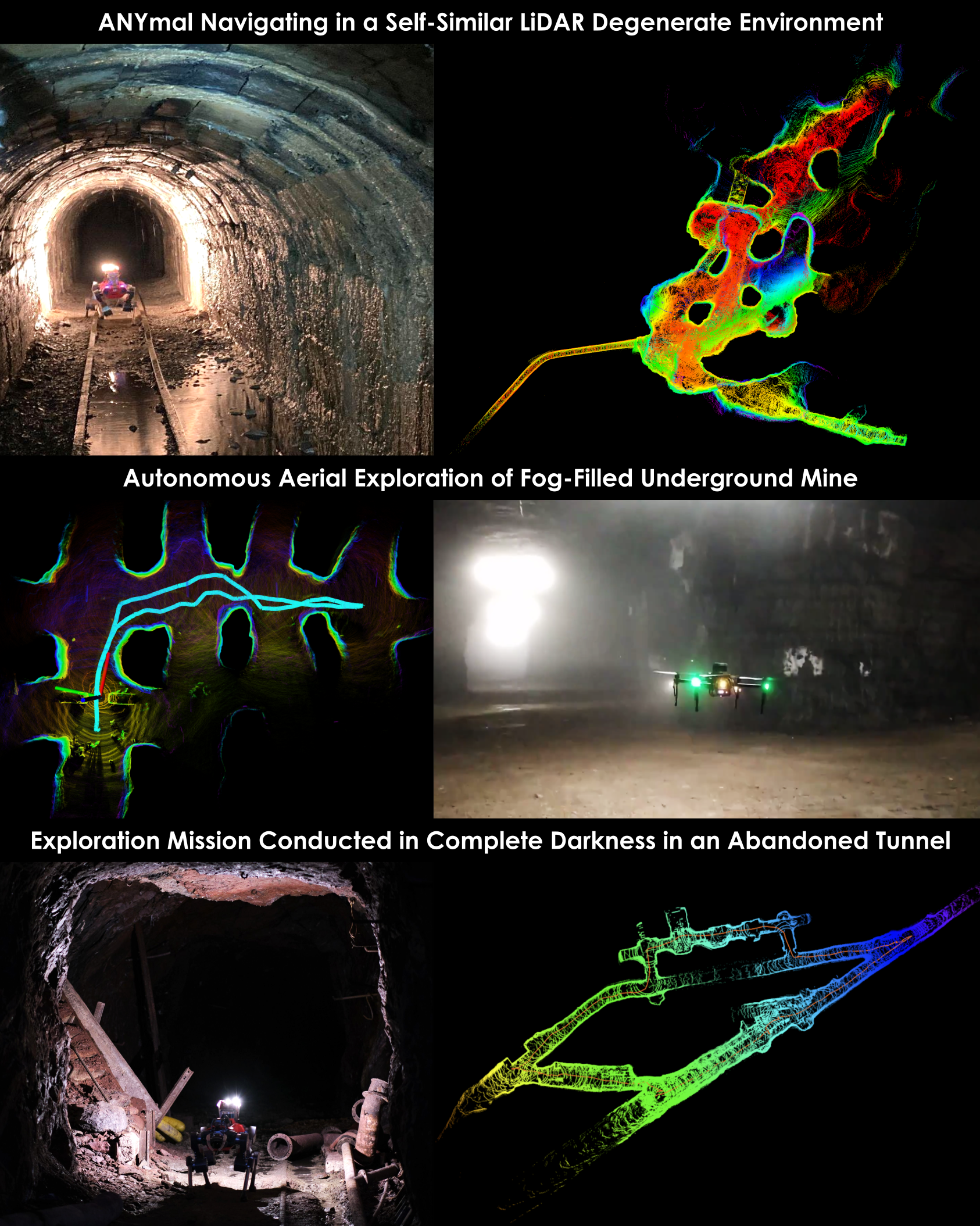}
	\caption{(Top) Autonomous exploration of a self-similar environment making LIDAR only localization unreliable. (Middle) Aerial exploration in a fog-filled environment using CompSLAM and exploiting thermal vision. (Bottom) 
	Underground tunnel environment exploration in conditions of darkness and subject to reflections due to mud/water puddles.
	\label{fig:cerberus:mapping} \vspace{-3mm}
	}
\end{figure}

\section{State of Practice and \\ Maturity of Underground SLAM\parlength{(9 pages)}}
\label{sec:maturity}
The previous section discussed state-of-the-art approaches for SLAM in underground environments across six \SubT teams. 
This section reports on the practical performance that can be achieved by these approaches, which provides 
useful data points to assess the maturity of \lidar-centric SLAM in underground worlds. 
We focus on three dimensions ---odometry, loop closures, and multi-robot mapping--- and for each we discuss performance and key aspects impacting it. 

\myParagraph{Odometry Estimation Accuracy}
This section shows that modern \lidar-centric odometry estimators can achieve a very low-drift (0.1-0.5\,\% of the trajectory traveled) in challenging underground environments.   
This enables impressive localization performance over long distances. 
For instance, \edited{\Cref{fig:CTU_CRAS_NORLAB_UAV_Localization_Error} shows results from
team CTU-CRAS-Norlab's unmanned aerial vehicles, achieving localization error under \SI{1}{\metre} in the Bull Rock cave system with flights reaching trajectory lengths of \SI{600}{\metre} and maximum velocities up to \SI{2}{\meter\per\second}}.

\emph{Multi-modality:}
Multi-modal sensing enhances robustness in challenging environmental conditions (\eg darkness, fog, smoke, dust, or feature-less scenes), as well as in the presence of hardware and software failures. 
\Cref{fig:cerberus:mapping} shows the map obtained by the multi-modal, onboard CompSLAM approach  by team CERBERUS; 
 CompSLAM achieves a low-drift trajectory estimate in extreme conditions with significant dust and obscurants. 
Although primarily driven by \lidar, CompSLAM also uses other modalities (\eg kinematic odometry or thermal) 
that are less sensitive to dense obscurants. This is still achieved on a modest computational budget: CompSLAM has been deployed on both ANYmal C robots that are equipped with powerful processors (i7-class systems), \edited{and} on the RMF-Owl aerial robot~\cite{de2022rmf}, that relies on a single-board computer.

\begin{figure}[b!]
\centering
	\includegraphics[width=1.0\columnwidth, trim= 0mm 0mm 0mm 7mm, clip]{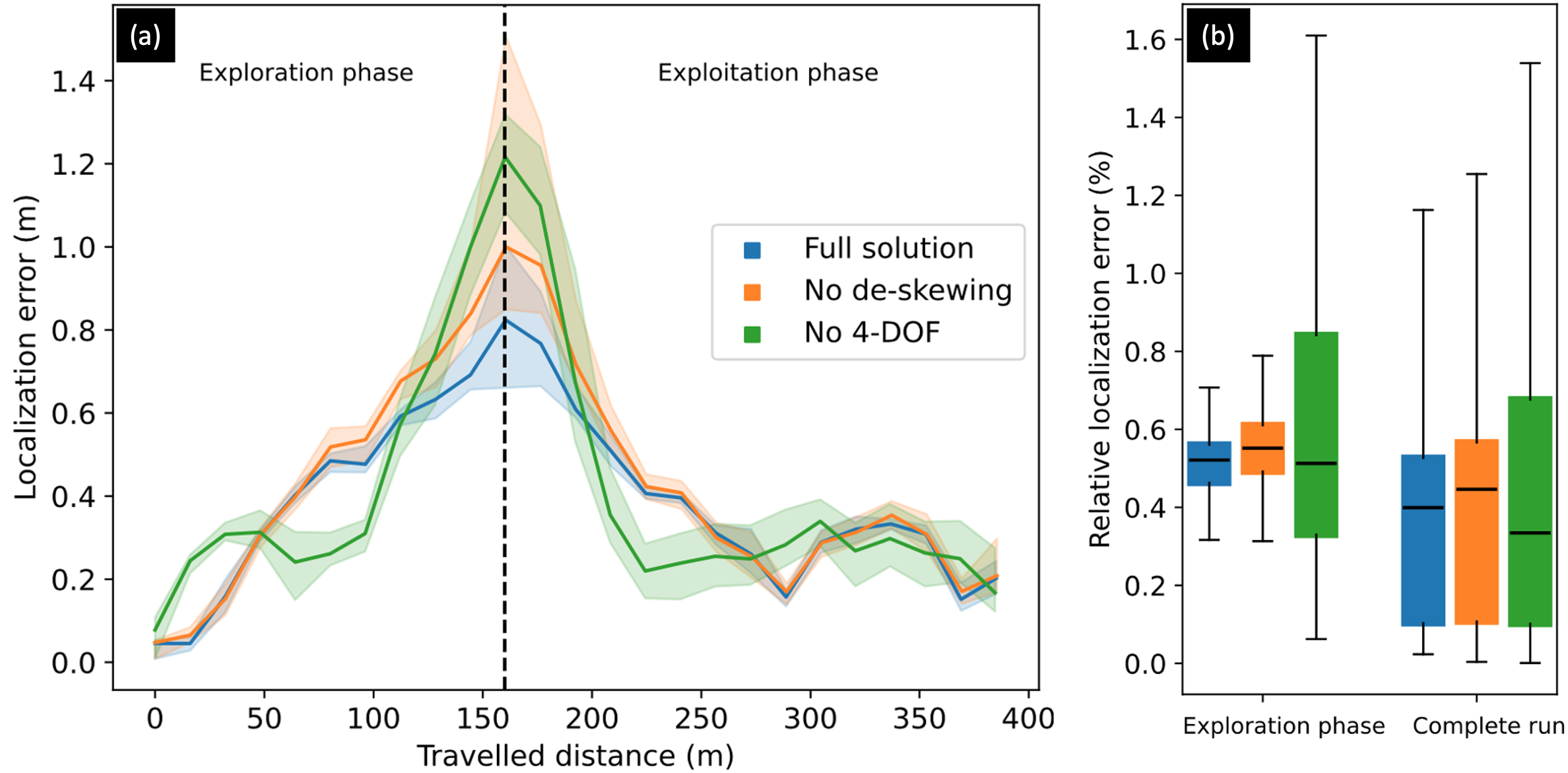}
	\caption{Localization error as a function of distance traveled. The solid lines are the median error, and the colored areas represent the first and third error quartiles. The dashed line delimits the exploration phase, during which the robot explored new areas, before returning to previously visited areas. 
	Statistics are computed over ten experiments.
	\label{fig:CTU-CRAS-Norlab-UGV-localization-error} \vspace{-3mm}
	}
\end{figure}

\begin{figure*}[b!]
    \centering
    \includegraphics[trim={0cm 0cm 0cm 0.4cm},clip, width=0.95\textwidth]{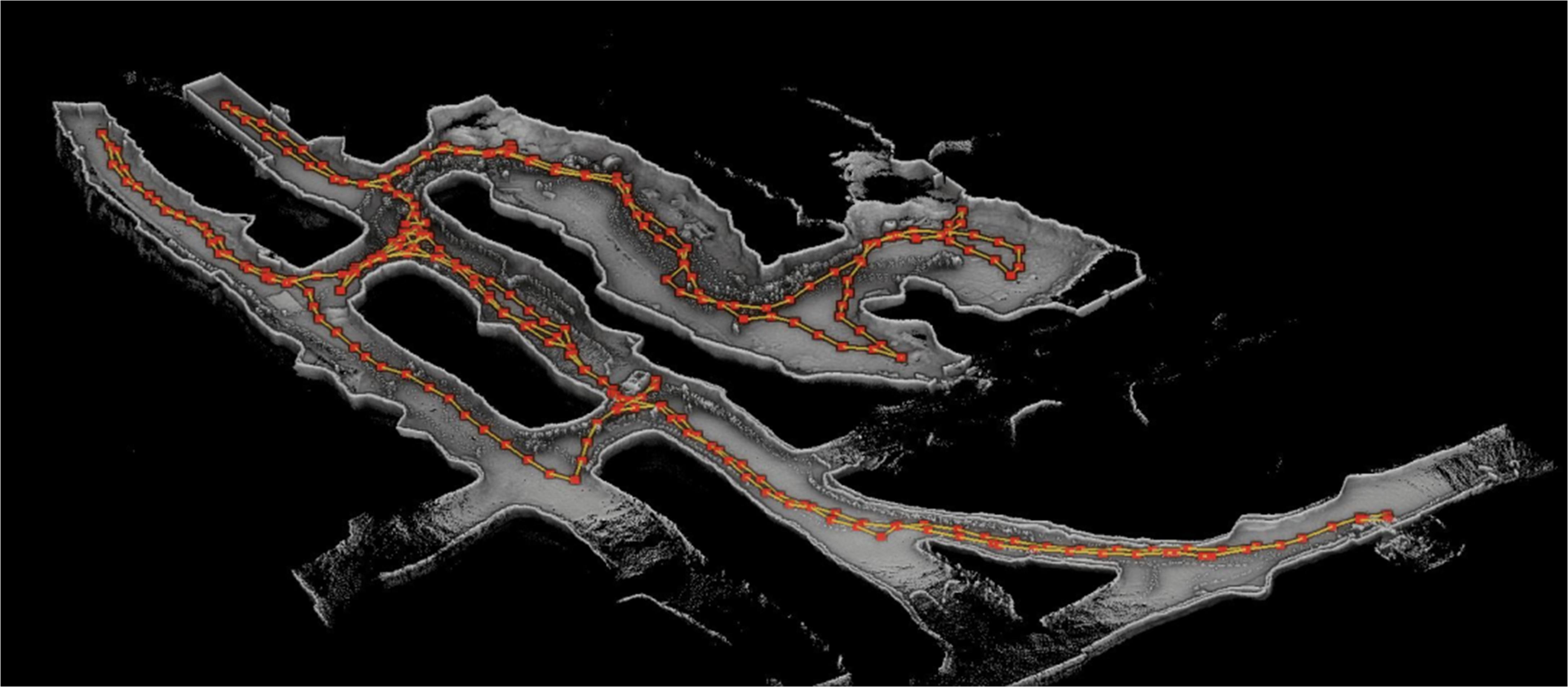}
    \caption{Team Explorer mapping results in Brady's Bend cave near Pittsburgh, PA, on a wheeled ground robot. The red dots represent the key poses and yellow edges show potential loop closure edges.
    \label{fig:cave} \vspace{-3mm}
    }
\end{figure*}
\begin{figure*}[b!]
\centering
	\includegraphics[width=1.95\columnwidth, trim= 0mm 0mm 0mm 0mm, clip]{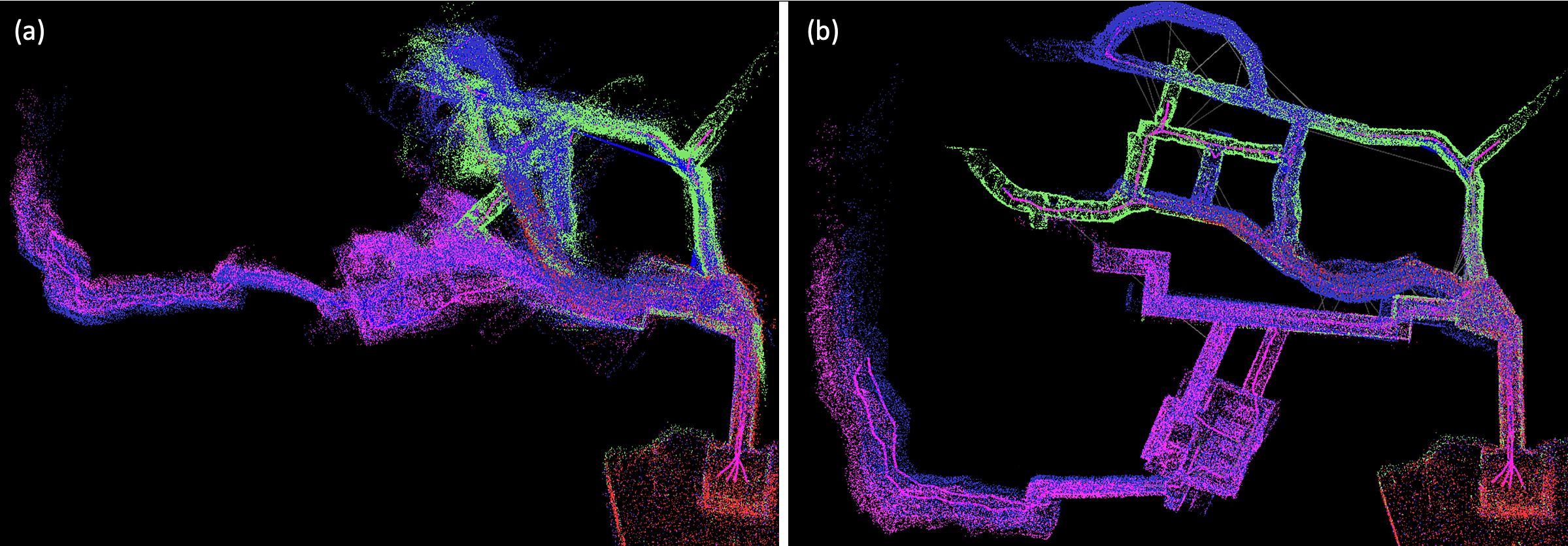}
	\caption{CoSTAR's SLAM results (a) without and (b) with the outlier rejection module during the preliminary run of the \SubT Finals. Each color represents the map of a different robot (four in total), blue lines represent accepted loop closures and gray lines represent rejected loop closures. GNC is able to successfully reject numerous false loop closures (gray lines in (b)).
	\label{fig:costar:gnc_prelim2}\vspace{-7mm} 
	}
\end{figure*}
\emph{\lidar pre-processing:}
\lidar data pre-processing is a key ingredient for accurate odometry estimation.
\Cref{fig:CTU-CRAS-Norlab-UGV-localization-error} shows an ablation study conducted by team CTU-CRAS-NORLAB on an unmanned ground vehicle, which highlights the impact of de-skewing the \lidar scans, as well as the impact of constraining the roll and pitch of the platform using IMU data during ICP (see~\Cref{sec:team-ccn}).
\edited{The path consists of a robot traveling through an unknown environment up to \SI{150}{\m} (the ``exploration'' phase), to which point it turned around to come back to the base station (the ``exploitation'' phase). Although CTU-CRAS-Norlab's SLAM solution does not use loop closures, it assumes low odometry drift and can reuse its global map for scan-to-map matching when revisiting known environments.}
All curves in \Cref{fig:CTU-CRAS-Norlab-UGV-localization-error}(a) exhibit 
{increasing errors (drift) during exploration}  but the de-skewing and roll-and-pitch-constrained optimization lead to reduced errors.
The result is confirmed by the localization error box plots in~\Cref{fig:CTU-CRAS-Norlab-UGV-localization-error}(b). \lidar pre-processing (\eg point down sampling via voxel grid filtering) is also crucial to reduce the computational burden, see the analysis in~\cite{andrzej2022iros}.

\myParagraph{Importance of Loop Closures} 
While \lidar-centric solutions compute low-drift odometric trajectories, such trajectory estimates keep accumulating error over time.
With a 0.5\% odometry drift, a robot would have \SI{5}{\metre} error after \SI{1}{km} traverse. This stresses the importance of detecting and enforcing loop closures to keep the localization error \emph{bounded}. \Cref{fig:cave} provides an example of accurate localization and mapping results by team Explorer, achieved by successful detection of loop closures. The figure shows mapping results in Brady's Bend cave near Pittsburgh, PA, on a wheeled ground robot. 
According to DARPA, team Explorer's SLAM system achieved a deviation of $6\%$ in the grand finale of the DARPA \SubT challenge, a performance that is the second best among all competing teams, behind CSIRO's Wildcat.\footnote{Team Explorer's performance was also recognized with the ``Most Sectors Explored Award'' by DARPA.}

\begin{figure}[t!]
\centering
	\includegraphics[width=1.0\columnwidth, trim= 0mm 0mm 0mm 1mm, clip]{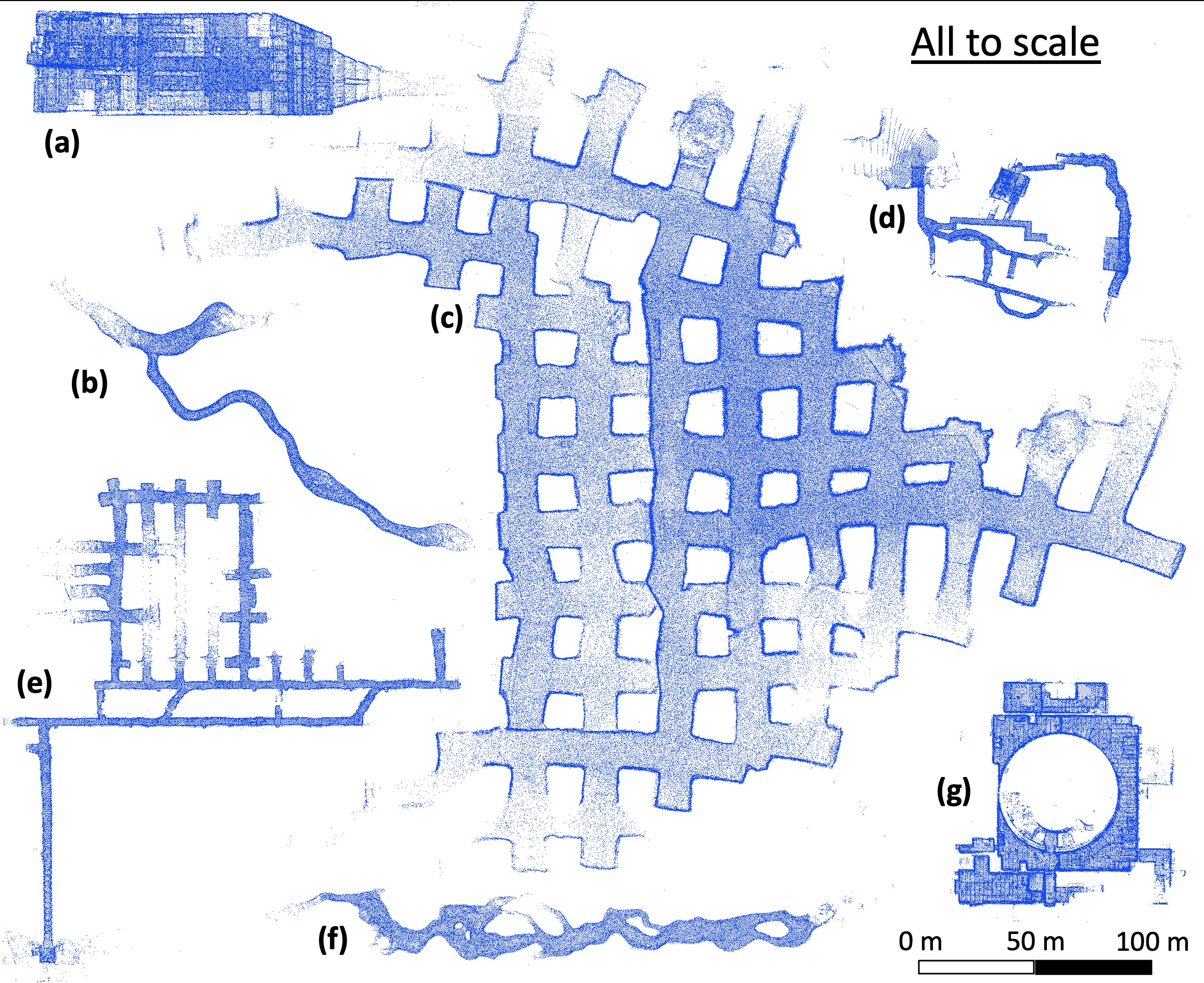}
	\caption{A to-scale representation of the robot-produced maps across a variety of environments team
	 CoSTAR tested in. (a) a 3-level abandoned subway in Los Angeles. (b) A lava tube in Lava Beds National Monument. (c) part of Kentucky Underground. (d) DARPA-created \SubT Finals course. (e) Bruceton Research Mine (\SubT Tunnel Competition). (f) Valentine cave (a lava tube) in Lava Beds National Monument. (g) Satsop power plant (\SubT Urban Competition). All maps are the best runs from a single-robot.
	\label{fig:costar:compositemaps} \vspace{-3mm}
	}
\end{figure}
\begin{figure}[t!]
 \centering
 \includegraphics[width=1.0\linewidth, trim= 30mm 50mm 30mm 50mm, clip]{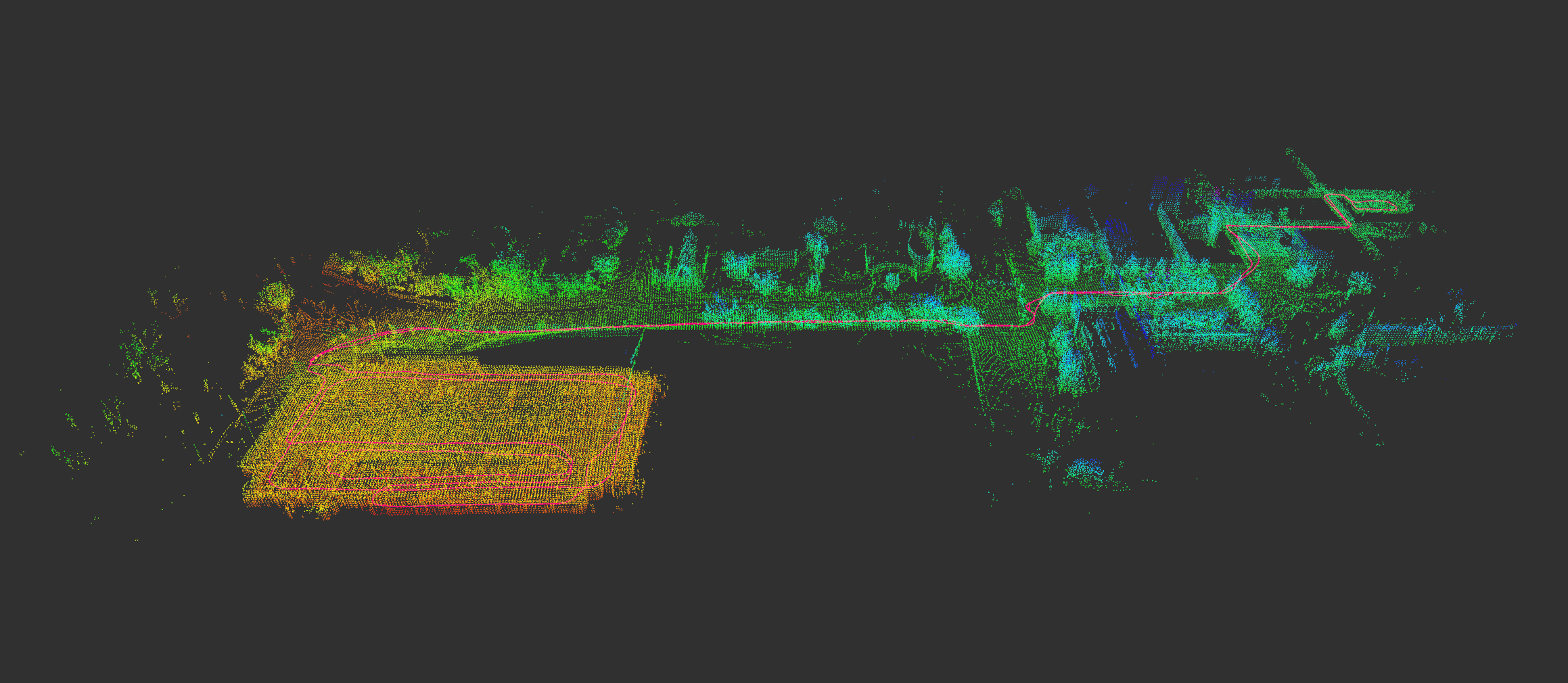}
 \caption{LIO-SAM map generated by MARBLE's Spot robot traversing from the University of Colorado-Boulder Engineering Center to the bottom of a nearby parking garage and back. The trajectory is marked in pink, and the point cloud is colored by elevation from red low to blue high.
\label{fig:cu_campus}\vspace{-3mm}
 }
\end{figure}

\emph{Robustness to outliers:}
\lidar-based loop closure detection is quite challenging in underground scenarios due to perceptual aliasing. At larger scales, and with more robots, the chances of false positive loop closures increase, especially in environments with self-similar locations. False loop closures, if not rejected, can have a negative impact on localization performance and lead to dramatic distortions in the map. \Cref{fig:costar:gnc_prelim2} shows CoSTAR's SLAM results (a) without and (b) with outlier rejection. CoSTAR's GNC-enabled~\cite{yang2020graduated} approach has been shown to produce accurate maps and reject up to 90\% outlier loop closures during the Final event of the DARPA \SubT challenge~\cite{kamak2022iros}. As depicted in~\Cref{fig:costar:compositemaps}, CoSTAR's outlier-robust loop closure detection enables creating high-precision 3D maps from multi-level urban environments \edited{with a combination of large rooms and small spaces}, to complex weaving lava tubes, to mines that are massive in scale, and finally the narrow passages found in the \SubT Final event.

\begin{figure}[ht!]
 \centering
 \includegraphics[width=1.0\linewidth]{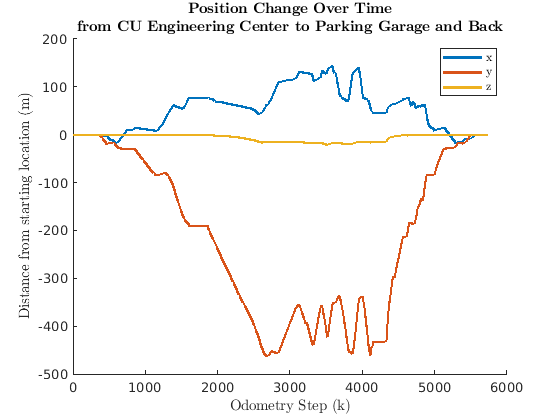}
 \caption{Position $x,y,z$ trajectory data of path in Figure \ref{fig:cu_campus}. The final position offset from the initial starting location was, \SI{0.31}{\metre} with a total trajectory length of \SI{2.2}{km}.
 \label{fig:marble_path}\vspace{-3mm}
 }
\end{figure}

\emph{Heterogeneous environments:}
Other examples of high precision localization and mapping in a large-scale and long-duration exploration are shown in~\Cref{fig:cu_campus} and \Cref{fig:marble_path} for the LIO-SAM system adopted by team MARBLE.
In these experiments, a robot is teleoped from within the University of Colorado-Boulder Engineering Center through all three levels of a parking garage before returning to its approximate original location in an hour-long operation. 
The test spans heterogeneous environment types, from tight urban indoor environments (with sharp turns, feature-less and narrow corridors, and staircases) to wide-open outdoor environments. The SLAM system accurately maintains elevation estimation through multiple levels of the parking garage with high level of geometric self-similarity while relying on only the OS1 \lidar and IMU. The \SI{2.2}{km} long trajectory shows a position difference of \SI{0.31}{\metre} from the start to the final position, which is equivalent to an error (after loop closures) of just $0.014\%$.

\begin{figure}[t!]
		\centering
		\includegraphics[height=0.5\textwidth]{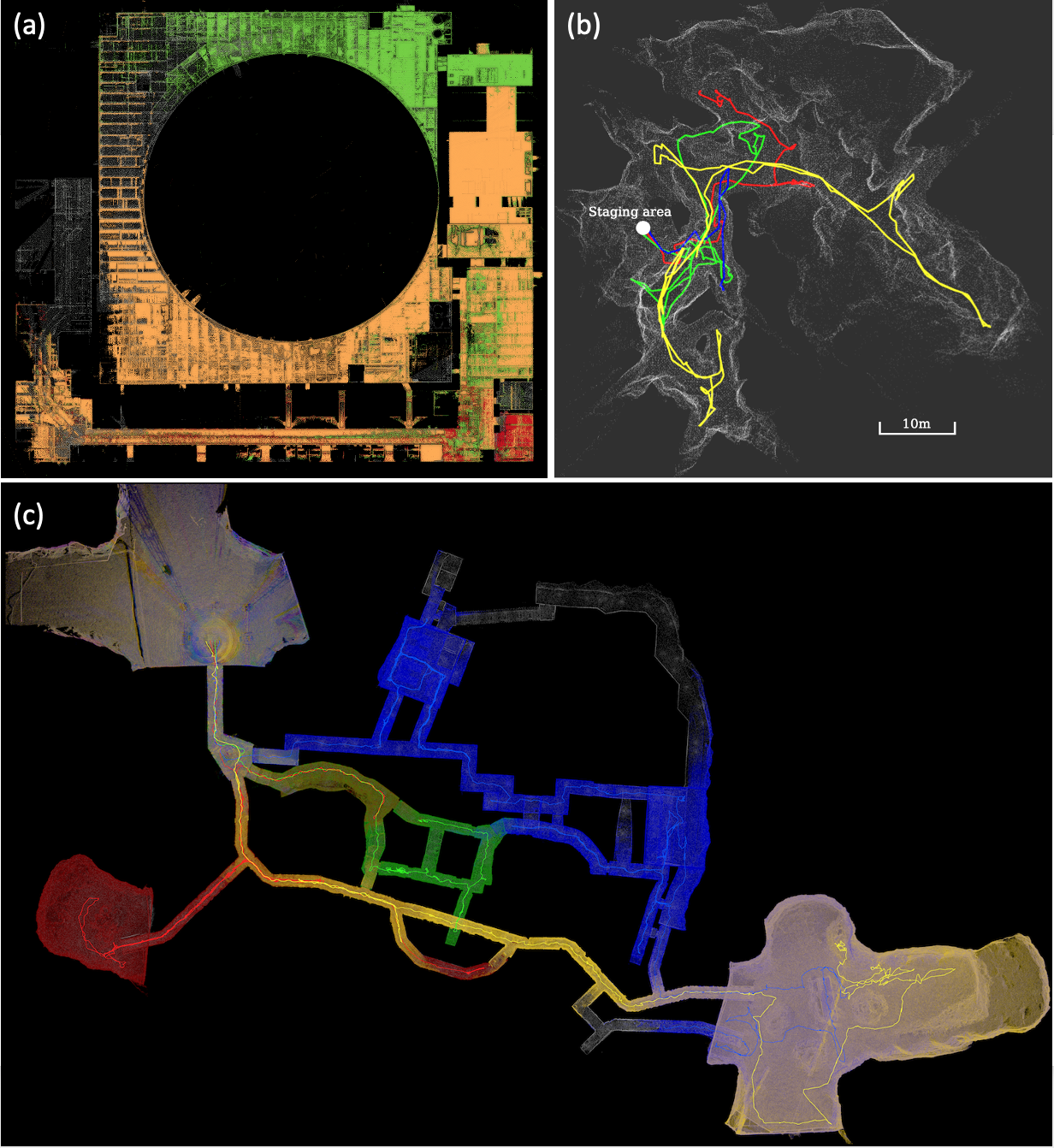}
		\caption{CSIRO's Wildcat results: (a) Point cloud map produced by three ground robots in the Beta Course of the Urban Event;
		estimated map is color-coded by agent, while DARPA reference point cloud is shown in gray. 
		(b) Map produced during cave testing at Capentaria Caves, Chillagoe, Queensland; 
		the merged agent-collected map is shown in gray, with agent trajectories in colors (drone colored as yellow). 
		(c) Merged point cloud map from the Final Event, color-coded by agent, with DARPA reference cloud in gray.
\label{fig:CSIROresults}\vspace{-3mm}
		}
\end{figure}

\myParagraph{Importance of Multi-robot Operation}
Multi-robot SLAM allows mapping larger areas while simultaneously reducing the localization and mapping errors thanks to inter-robot loop closures.
\Cref{fig:CSIROresults} shows the maps produced by CSIRO's Wildcat decentralized multi-robot SLAM  system in two \SubT events (Urban and Final) and in a cave in Australia. The map in~\Cref{fig:CSIROresults}(a) is built by three ground robots, while the maps in~\Cref{fig:CSIROresults}(b-c) are created by four robots (including a UAV, in the cave case). 
\edited{According to DARPA, in the Final \SubT event Wildcat produced the top map with less than $1$\% deviation from the ground truth where
they defined deviation as the percentage of points in
the submitted point cloud that are farther than one meter from
the points in the surveyed point cloud map. Wildcat also produced the single  most accurate reports in the Urban and Final events with \SI{22}{~cm} and \SI{4.8}{~cm}
 error, respectively.}
We refer the \edited{reader to~\cite{Wildcat}} for a more extensive experimental evaluation. Additional qualitative results produced by Wildcat in perceptually challenging environments are also available on the websites of two commercial partners of CSIRO, Emesent \cite{emesent} and Automap \cite{automap}.

\begin{figure}[t]
\centering
	\includegraphics[width=1.0\columnwidth, trim= 0mm 0mm 0mm 0mm, clip]{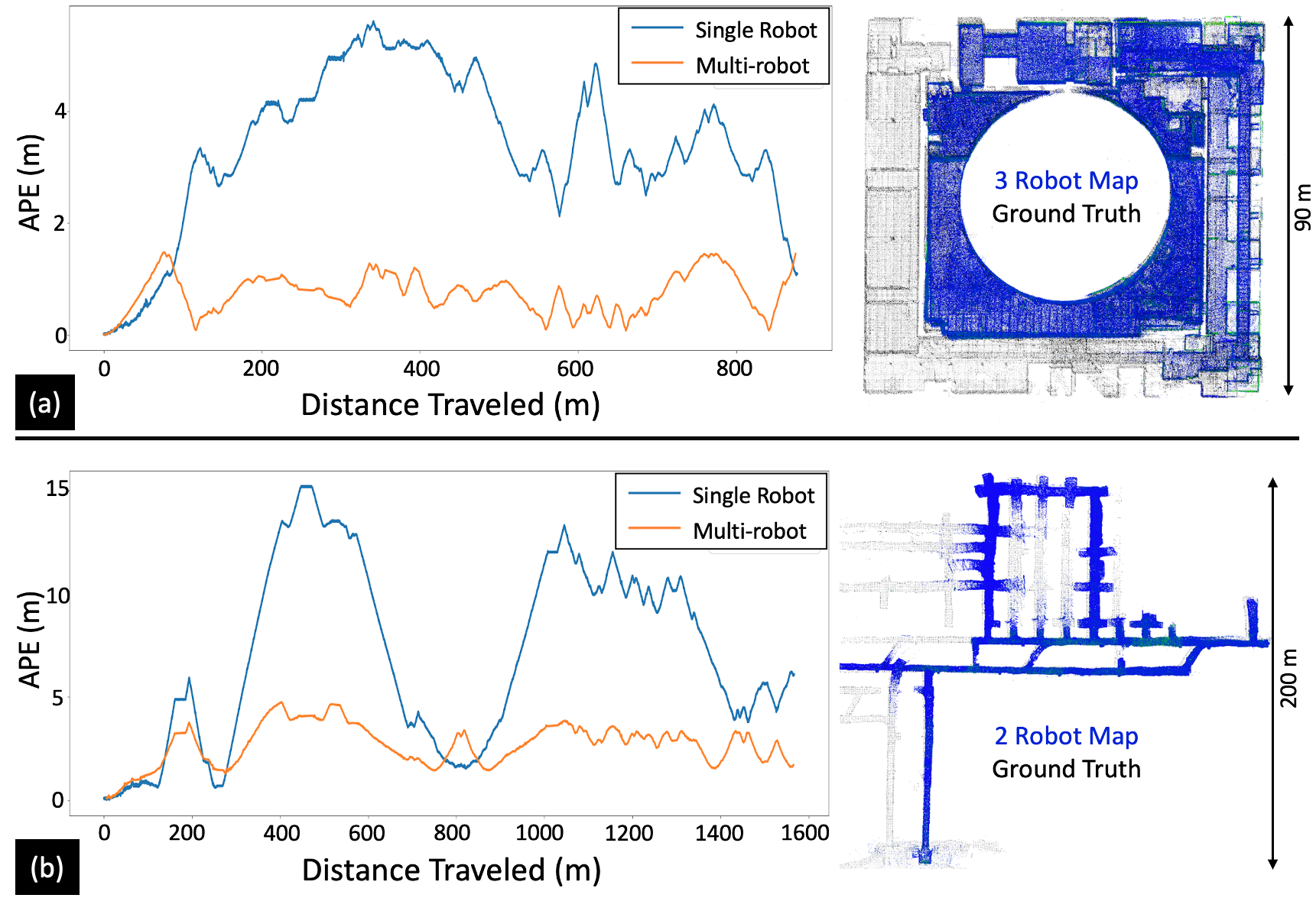}
	\caption{CoSTAR results: improvement in Absolute Pose Error (APE) due to inter-robot loop closures and multi-robot pose graph optimization, with the resulting multi-robot map on the right. (a) Test with three robots in the Satsop power plant. 
	(b) Test with two robots in the Bruceton Research Mine.
	\label{fig:costar:multi-rob-improve} \vspace{-5mm}
	}
\end{figure}

\emph{Inter-robot loop closures:}
\Cref{fig:costar:multi-rob-improve} shows the dramatic reduction of the Absolute Pose Error (APE) in team CoSTAR's SLAM architecture due to inter-robot loop closures and multi-robot pose graph optimization. As in the single-robot case, capitalizing on inter-robot loop closures requires a good strategy for outlier rejection, since many inter-robot loop closure detections will be incorrect due to perceptual aliasing.

\begin{figure}[t]
\centering
\includegraphics[trim={0cm 0cm 0cm 0cm},clip, width=1.0\linewidth]{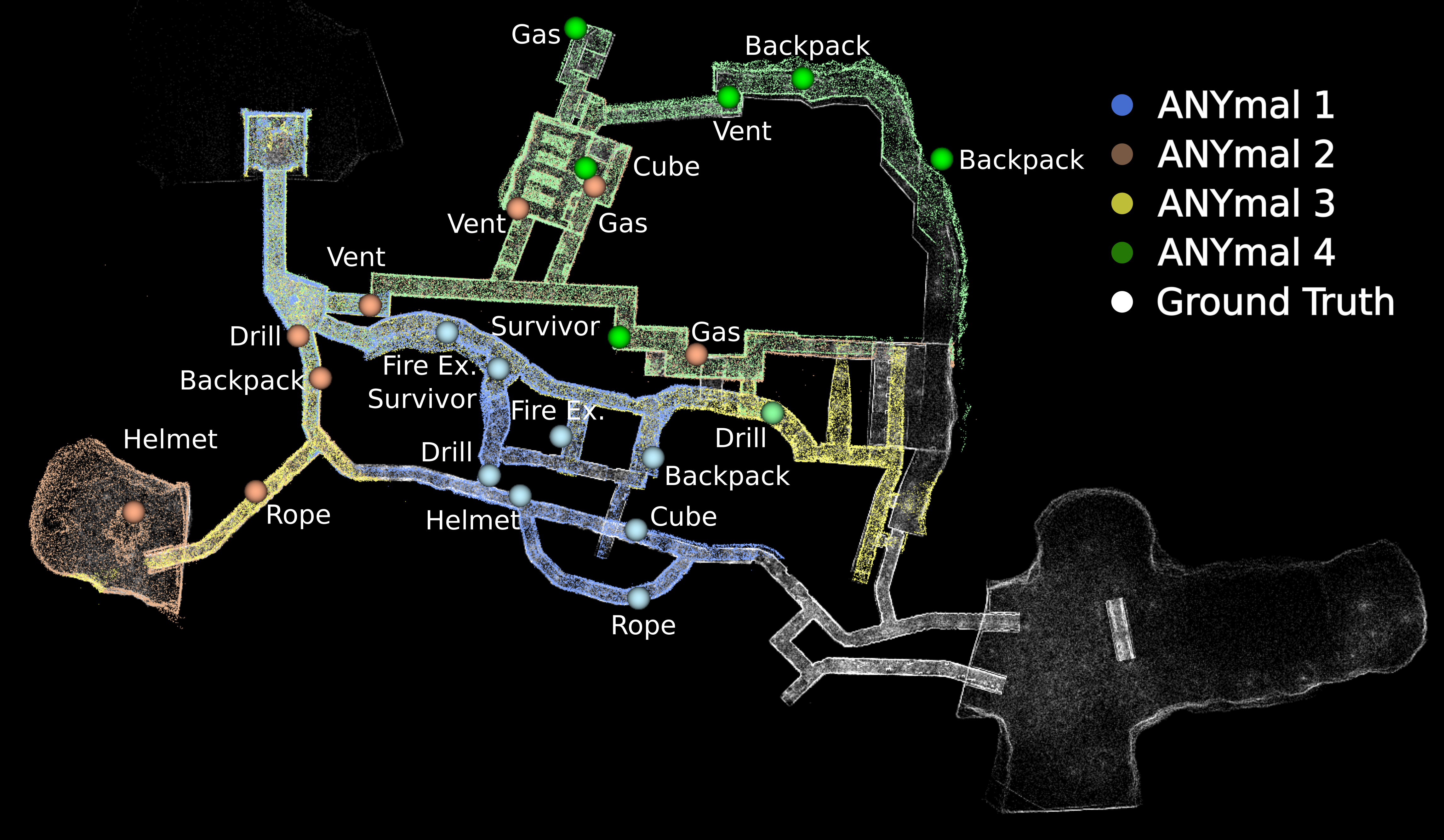}
\caption{CERBERUS results: Onboard maps for all deployed robots generated using CompSLAM are overlaid on top of the DARPA provided ground truth for the final event of the DARPA SubT Challenge. The scored artifacts are shown and are colored to correspond with the reporting robot.}
\label{fig:cerberus:multi_robot_subt} 
\vspace{-3mm}
\end{figure}

\edited{In the DARPA \SubT finals event, team CERBERUS deployed four ANYmal quadrupedal robots to autonomously navigate a total distance of \SI{1.75}{\km}. 
The maps generated by the onboard solution, CompSLAM, along with scoring artifacts, are qualitatively compared against the DARPA-provided ground truth map in~\Cref{fig:cerberus:multi_robot_subt}. The individual robot results are made globally consistent by M3RM by exploiting inter-robot loop closures; a quantitative comparison between the onboard and global mapping approaches is presented in~\Cref{tab:prize_run_evaluation_ape}.}

\begin{table}[h!]
\centering
\setlength{\tabcolsep}{4pt}
\begin{tabular}{c|cc|cc}
\toprule
\textbf{Robot} & \multicolumn{2}{c|}{\textbf{CompSLAM (Onboard)}} & \multicolumn{2}{c}{\textbf{M3RM (Server)}}\\
\midrule
& Rotation\,[$\degree$] & Translation\,[\si{\metre}] & Rotation\,[$\degree$] & Translation\,[\si{\metre}]\\
ANYmal 1 & 2.45\,(0.67) & 0.72\,(0.41) & \textbf{1.59}\,(0.46) & \textbf{0.25}\,(0.13) \\
ANYmal 2 & 3.97\,(0.40) & 1.29\,(0.90) & \textbf{0.96}\,(0.30) & \textbf{0.36}\,(0.28) \\
ANYmal 3 & \textbf{0.89}\,(0.50) & 0.23\,(0.43) & 2.30\,(1.02) & \textbf{0.20}\,(0.34) \\
ANYmal 4 & 2.22\,(0.79) & 1.00\,(0.71) & \textbf{2.16}\,(0.55) & \textbf{0.24}\,(0.17) \\
\bottomrule
\end{tabular}
\caption{Comparison of the mean and standard deviation of the Absolute Pose Error (APE) for CERBERUS' CompSLAM (each robot) and M3RM (all robots considered together) approaches for the DARPA SubT challenge Final Event.}
\label{tab:prize_run_evaluation_ape} 
\vspace{-3mm}
\end{table}

\begin{figure}[t!]
    \centering
    \includegraphics[trim={0cm 0cm 0cm 0cm},clip, width=1.0\linewidth]{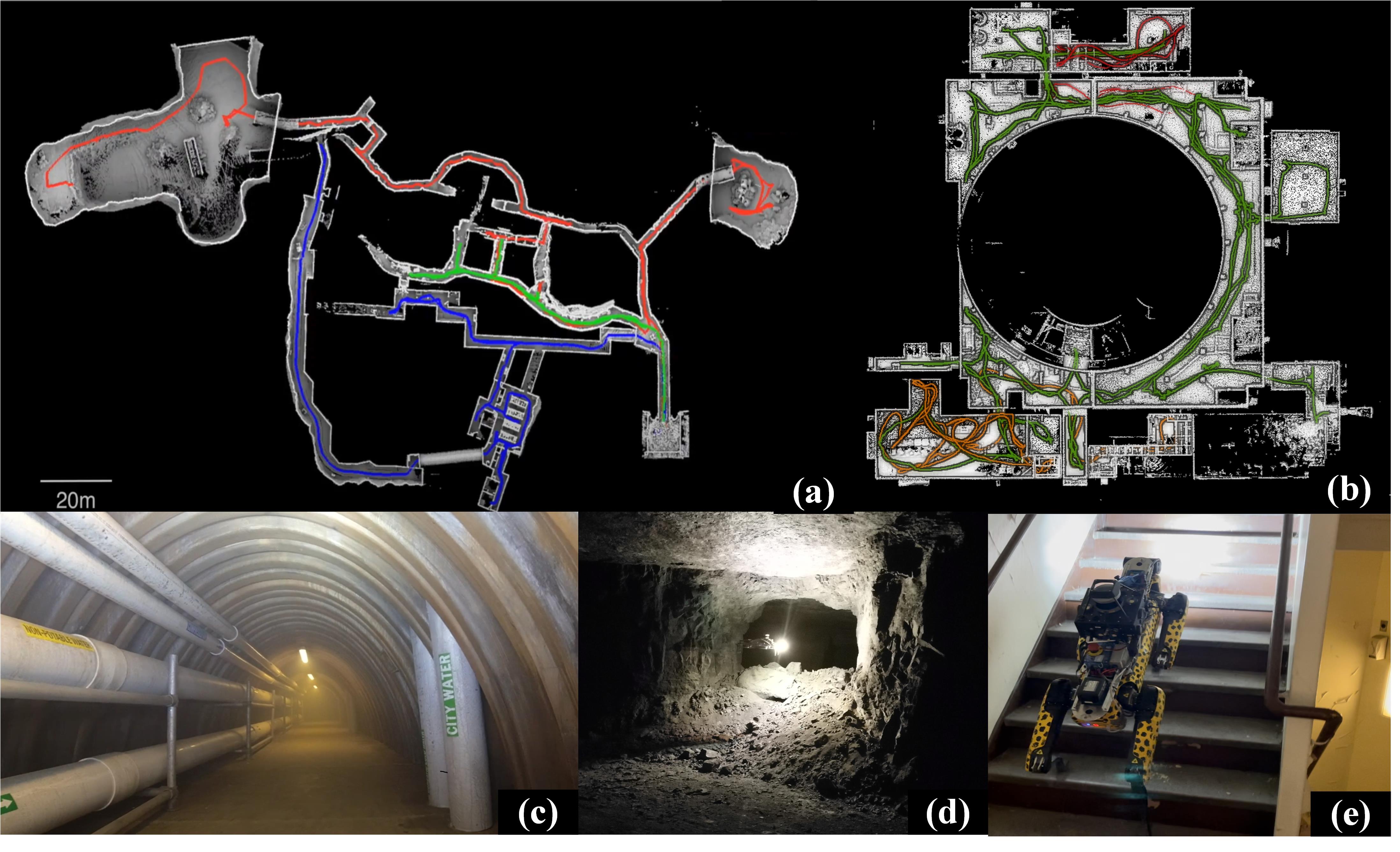} \vspace{-3mm}
    \caption{Testing sites and results of Explorer's SLAM system. (a) UAV, UGV, and legged robot exploration in the Final Circuit of \SubT, including urban, cave, and tunnel environments. (b) UAV and UGV exploration in 
    the Urban Circuit of \SubT. (c) Tunnel environments, test site with smoke. 
    (d) Cave environment UAV test site. 
    (e) Urban environment with legged robot.
    \label{fig:explorer_darpa} \vspace{-3mm}
    }
\end{figure}

\emph{Heterogeneous teams:}
We already commented on the benefit of having heterogeneous sensing capabilities. Here, we discuss the advantage of using heterogeneous platforms for exploration. Indeed, most of the \SubT teams used a combination of wheeled and legged ground robots and UAVs.
\Cref{fig:explorer_darpa}(c) shows Explorer's mapping result in the Urban Challenge Alpha Course reconstructed by multiple robots (UGV1, UAV1, and UAV2) operating in a dark and foggy environment with a vertical shaft. 
Green, orange, and red lines are the estimated trajectories of UGV1, UAV1, and UAV2 respectively. 
\Cref{fig:explorer_darpa}(a) shows the mapping result in the \SubT Finals by a heterogeneous fleet. The blue, green and red lines are the estimated trajectories of UGV1, UGV2, and UGV3 respectively. The travel distance of UGV1, UGV2, and UGV3 are \SI{445.2}{\metre}, \SI{499.8}{\meter}, and \SI{596.6}{\meter}, respectively. 
Explorer's SLAM solution achieved accurate localization and mapping despite the challenging environmental conditions, including low light, long corridors, heavy dust/fog, and even dynamic scenes. Heterogeneity enables mapping a broader variety of environments (\eg UAVs enable exploring 
vertical shafts) and allows richer exploration strategies (\eg using UAVs for fast exploration, and UGVs for more accurate mapping).

\bib


\section{Future Research Directions \\ and  Open Problems\parlength{(3 pages)}}
\label{sec:openProblems}

In the light of the results in~\Cref{sec:maturity} and the outcome of the DARPA \SubT competition, 
this section provides a summary of which problems in underground SLAM can be considered solved \edited{or can be solved with some good engineering} and what are still open problems that will likely require more fundamental research.

\lidar-centric SLAM solutions have become increasingly robust to challenging environments. 
Feature detection or scan pre-processing enable real-time point cloud alignment. 
Tight coupling with inertial data enables more robust motion estimation, 
by allowing de-skewing the \lidar scans, bootstrapping ICP-based scan matching, and potentially eliminating roll and pitch drift.
Keyframe-based or sub-map-based approaches, combined with a factor graph framework, allow sparsifying the trajectory into a reduced set of poses and enable online operation in large-scale, long-term, multi-robot explorations with reduced computational complexity. The addition of other sensing modalities further increases robustness.

Looking across the six solutions examined in~\Cref{sec:architectures}-\ref{sec:maturity}, there is reason to believe that the underground SLAM problem, with high-quality multi-modal sensing suites, is a solved problem. Yet only solved with sufficient qualifications of the environment, the scale, the sensors, the parameter tuning, and the computation power. We believe that in the context of extreme subterranean environments, the majority of open problems defined in~\cite{cadena2016slam} still applies. In the rest of this section, we highlight current challenges and open problems in underground localization and mapping.

\myParagraph{Robust and Resilient Perception}
One of the common failure modes observed across most of the presented architectures is localization failure due to falls, drops, or collisions~\cite{Vimo} when traversing rough terrains in unstructured underground environments. These-high frequency motions are not entirely captured by the onboard perception system, \eg due to the lower sampling frequency of structured-light sensors~\cite{Khader}. This could lead to poor motion estimates and eventually localization failure. With robotic systems that can withstand a fall and continue to operate (\eg Boston Dynamics Spot, Flyability drones, BIA5 Titan, \edited{ANYmal, RMF-Owl}), a relatively under-explored area is reliable state estimation under unexpected collisions and temporary interruptions of the sensor streams.
\edited{Although early work on localization subject to collision shows promising results~\cite{Deschenes2021}, better exploring the limitations of different systems and algorithms in “crash tests” scenarios would help improve all-round real-world robustness.} Furthermore, engineering work in incorporating velocity-based sensors (\eg event-based cameras~\cite{eventcameras}) which might maintain ego-motion tracking without saturation during adverse events could greatly benefit SLAM systems.

\edited{At a more fundamental level, underground operation requires redundancy and resourcefulness, but this needs to be achieved beyond just ``adding more sensors''.} The SLAM literature is lacking fundamental research in \emph{resilient} algorithms and systems. While robust systems are designed to withstand (often small) disturbances (\eg degraded sensing or environmental changes), resilient methods dynamically reconfigure to regain performance in the face of changing environmental stressors~\cite{Prorok22axiv-resilience}. For instance, a resilient system would dynamically change its parameters (or even its algorithmic components) depending on the scenario, contrarily to the current SLAM systems, which are ``rigid'' and heavily rely on manual parameter tuning; see the comments about parameter tuning in the dirty details subsections in \Cref{sec:architectures}, as well as the discussion about the ``curse of parameter tuning'' in~\cite{cadena2016slam}.

\myParagraph{\edited{Beyond Traditional SLAM Sensors}}
Achieving robustness under perceptual aliasing, dense obscurants, and severe environment degradation remains a challenge and can benefit from 
incorporating non-traditional sensing modalities and designing methods for failure detection and recovery. \edited{Thermal vision allows penetrating conditions of visual degradation, where cameras and \lidars fail due to the presence of obscurants. Similarly, radar} is able to maintain  localization despite the presence of fog, as the wavelengths in commercial automotive millimeter-wave radars are large enough to bypass particulate such as fog and dust that causes spurious reflections that render \lidar point clouds unusable for localization and mapping purposes. While research into millimeter-wave radar-based localization~\cite{kramer2020radara,lu2020see,kramer2020radarb,lu2020milliego} and the creation of radar factors for SLAM applications is ongoing, \edited{including the release of public datasets such as~\cite{kramer2021coloradar,kim2020mulran}}, the integration of these sensors is not as established as other sensing modalities, due to complexity of the corresponding sensor models and data association. Multi-modal SLAM systems could \edited{also be} pushed further by developing failure detection and recovery methods. 
\edited{Autonomous exploration of subterranean settings requires dynamically adaptive algorithmic architectures to achieve solution resourcefulness.} 
Still related to resilient operation, it would be desirable to design approaches that can detect failures of a sensing modality and reconfigure the system accordingly. The importance of degeneracy detection in multi-modal sensing is discussed in~\cite{ebadi2021dare}, while fault detection in perception system is investigated in~\cite{Antonante22arxiv-perceptionMonitoring}.

\myParagraph{Scaling Up: Centralized vs. Distributed Systems}
Multi-robot \lidar-centric SLAM is a mature research area. This paper showed that centralized approaches can achieve accurate and real-time performance for moderate team sizes (5-10 robots); moreover, decentralized approaches attain small errors even without relying on inter-robot loop closures in moderate-scale scenarios (\eg $<$\SI{1}{km} traversal). 
However, scaling up SLAM solutions to very large teams (\eg $>$100 robots) and very large-scale scenarios \edited{(\eg city-scale~\cite{AutoMerge} and forest-scale~\cite{Baril2022})} is likely to require a more distributed approach. 
In centralized approaches, large team sizes would quickly reach a bottleneck in terms of communication as well as processing at the base station.\footnote{The \SubT teams carefully handled communication (\eg via compression and down-sampling) to meet the bandwidth constraints. Moreover, teams using centralized solutions relied on powerful base stations, \eg CoSTAR' base station relied on an AMD Ryzen Threadripper 3990X with 64 cores/128 threads at 2.9GHz.} 
Therefore, distributed architectures are likely to be needed to scale up operation. For large fleets covering large-scale geographic areas, it will be necessary to consider (i)~resource-aware collaborative inter-robot loop closure detection techniques~\cite{tian2021resource,giamou2018talk,tian2018near} that intelligently utilize limited mission-critical resources available onboard (\eg compute, battery, and bandwidth) and 
(ii)~distributed factor-graph and pose-graph optimization methods~\cite{tian2021distributed,chang2021kimera,Tian21tro-KimeraMulti,futureMapping2}, both of which are active research areas.  
We also believe that hierarchical map representations (\eg~\cite{Hughes22rss-hydra}) will be needed for large-scale environments where point-cloud or voxel-based representations would clash with memory constraints. 

In terms of engineering, it would be desirable to develop and release open-source implementations of multi-robot SLAM systems.
As we observed, SLAM progress in \SubT was also enabled by the availability of high-quality open-source implementation for SLAM components (\eg the back-end provided by GTSAM) or entire systems (\eg LIO-SAM).
Therefore, the development of distributed SLAM systems will benefit from a similar open-source infrastructure. 

\myParagraph{Scaling Down: Miniaturization and Low-Cost Sensing}
All solutions examined in this paper leverage one or multiple \lidars, and powerful embedded computers. More work is required to enable the capabilities presented in this paper, but with low-cost components that might be suitable for smaller, cheaper, expendable systems. For instance, it would be desirable to deploy a large number of  expendable robots for high-risk missions (\eg search \& rescue, planetary exploration), or to design more affordable robots to increase adoption by first responders.  These platforms would ideally have a small form factor to enable exploration of narrow passages (\eg pipes) while being easy to transport by human operators.
Achieving this goal entails both engineering efforts (\eg development of novel sensors or specialized ASICs for on-chip SLAM~\cite{Navion}) and more research on vision-based SLAM in degraded perceptual conditions (\eg dust or fog). 
\bib


\section{Conclusion}
While progress in SLAM research has been reviewed in prior works, none of the previous surveys focus on underground SLAM. Given the astonishing progress over the past several years, this paper provided a survey of the state-of-the-art, and the state-of-the-practice in SLAM in extreme subterranean environments and reports on what can be considered solved problems, what can be solved with some good systems engineering, and what is yet to be solved and likely requires further research. We reviewed  algorithms, architectures, and systems adopted by six teams that participated in the DARPA Subterranean (\SubT) Challenge, with particular emphasis on \lidar-centric SLAM solutions, heterogeneous multi-robot operation (including both aerial and ground robots), and real-world underground operation (from the presence of obscurants to the need to handle tight computational constraints).
Furthermore, we provided a table of open-source SLAM implementations and datasets that have been produced during the 
\SubT challenge and related efforts, and constitute a useful resource for researchers and practitioners.

\bib

\bibliographystyle{IEEEtran}
\footnotesize
\bibliography{bibliography.bib}

\end{document}